\documentclass{article}

\usepackage{arxiv}
\usepackage[utf8]{inputenc} % allow utf-8 input
\usepackage[T1]{fontenc}    % use 8-bit T1 fonts
\usepackage{hyperref}       % hyperlinks
\usepackage{url}            % simple URL typesetting
\usepackage{booktabs}       % professional-quality tables
\usepackage{amsfonts}       % blackboard math symbols
\usepackage{nicefrac}       % compact symbols for 1/2, etc.
\usepackage{microtype}      % microtypography
\usepackage{graphicx}
\usepackage{natbib}
\usepackage{doi}
\usepackage{amsmath,amssymb,amsthm}
\usepackage{array}
\usepackage{float}
\usepackage{multirow}
\usepackage[table]{xcolor}
\usepackage{etoolbox}
\AtBeginEnvironment{table}{\small}
\definecolor{highlightgreen}{HTML}{9AD153}
\usepackage{pgfplots}
\pgfplotsset{compat=1.18}
\newtheorem{theorem}{Theorem}
\newtheorem{corollary}{Corollary}
\newtheorem{lemma}{Lemma}
\newtheorem{definition}{Definition}
\usepackage[capitalise]{cleveref}
\title{New Synchronous Computation Dynamics for Hopfield Networks}

\date{July 24, 2026}

\author{
	\href{https://orcid.org/0009-0008-8972-7260}{\includegraphics[scale=0.06]{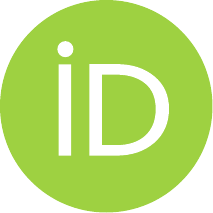}\hspace{1mm}Francisco Requena-Dom\'inguez} \\
	Department of Computer Languages and Computer Science \\
	University of M\'alaga \\
	Bulevar Louis Pasteur 35, 29071 M\'alaga, Spain \\
	\texttt{paco.requena@gmail.com} \\
    \And
	\href{https://orcid.org/0000-0001-8500-0488}{\includegraphics[scale=0.06]{orcid.pdf}\hspace{1mm}Rafaela Ben\'itez-Rochel} \\
	Department of Computer Languages and Computer Science \\
	University of M\'alaga \\
	Bulevar Louis Pasteur 35, 29071 M\'alaga, Spain \\
	\texttt{rbenitezr@uma.es} \\    
    \And
	\href{https://orcid.org/0000-0001-8231-5687}{\includegraphics[scale=0.06]{orcid.pdf}\hspace{1mm}Ezequiel L\'opez-Rubio} \\
	Department of Computer Languages and Computer Science \\
	University of M\'alaga \\
	Bulevar Louis Pasteur 35, 29071 M\'alaga, Spain \\
	and ITIS Software, University of M\'alaga \\
	C/ Arquitecto Francisco Pe\~nalosa 18, 29010 M\'alaga, Spain \\
	\texttt{ezeqlr@lcc.uma.es} \\
}

\renewcommand{\headeright}{Technical Report} %------------??????

\hypersetup{
pdftitle={New Synchronous Computation Dynamics for Hopfield Networks},
pdfsubject={Computational Dynamics of Neural Networks},
pdfauthor={Francisco Requena-Domínguez, Rafaela Benitez-Rochel, Ezequiel López-Rubio},
pdfkeywords={Discrete Hopfield Networks, handling tools, synchronous dynamics, convergence speed up, combinatorial optimization.},
}

\begin{document}
\maketitle

\begin{abstract}
    The dynamics of the original Hopfield network is asynchronous (sequential) (updates the state of only one neuron per time step). In this paper, we propose a new tool and a new dynamics to reduce the processing time by updating one or more neurons simultaneously per instant while ensuring process convergence and aiming for the maximum energy decrease at each step, thus guaranteeing the shortest total processing time.\\
From the point of view of synchronous dynamics, calculating the next network state at which energy decreases the most from the current state while ensuring convergence is itself a combinatorial optimization problem. We develop and use a new tool to solve it. We call this new tool \textbf{Discrete Differential Filter (DDF)} and, based upon it, we develop a new synchronous dynamics which we call \textbf{SD-DDF} (Synchronous Dynamics based upon Discrete Differential Filter). 
\\
In this paper, we review the original asynchronous dynamics for Hopfield networks and present a new tool and a new synchronous dynamics with its theoretical justification and four computational experiments to assess the speed up in processing time empirically.
\end{abstract}


\section{Introduction}
In 1982, the Hopfield network was introduced as a model for associative memory retrieval \cite{Hopfield82}, providing a framework in which information is stored in the form of stable equilibrium states of a dynamical system, where each stored pattern corresponds to a local minimum of an energy function. When the network is presented with a corrupted or incomplete version of a learned pattern, the collective dynamics guide the system toward the nearest energy minimum, effectively retrieving the original memory. Its ability to retrieve content-addressable memories and its rigorous energy-based formulation made it a foundational model in the study of recurrent neural systems and combinatorial optimization problems. However, the classical Hopfield network exhibits several well-known limitations. The storage capacity is relatively low, and the network may converge to spurious attractors or local minima that are not related to any trained pattern. Moreover, the conventional asynchronous update rule, which ensures the monotonic decrease of the energy function and guarantees convergence, imposes a sequential computation constraint that limits parallel execution and efficiency in large-scale implementations.
Recent years have witnessed a renewed interest in Hopfield-type models, driven by advances in energy-based learning, dense associative memories, and their reinterpretation as attention mechanisms in modern deep learning architectures. Despite this, most existing work has focused on improving memory capacity, robustness, or representational power through modified energy functions or learning rules. In contrast, the computational dynamics of Hopfield networks and, in particular,  the design of efficient parallel update schemes with theoretical guarantees, have received limited attention.
Addressing this gap is crucial, since a well-founded parallel update rule facilitates practical implementations on modern hardware, allowing energy-based associative memories to be more naturally integrated into contemporary machine learning and optimization processes. Furthermore, enabling parallel dynamics directly improves computational efficiency, making Hopfield-type models more suitable for large-scale problems and real-time applications.
In this paper, we first provide a concise review of the original sequential dynamics of the classical Hopfield network and the theoretical arguments that ensure its convergence. We then propose novel parallel dynamics as an alternative to the traditional sequential update rule. The proposed dynamics are supported by a theoretical analysis of their stability and energy behaviour. Finally, we present computational experiments that empirically demonstrate the speedup achieved by the proposed parallel dynamics compared to the classical sequential approach, highlighting its practical advantages without compromising solution quality.

\section{Related Works}
The original Hopfield network introduced an asynchronous update mechanism that guarantees a monotonic decrease in the energy function and convergence towards fixed-point attractors. This property established the model as a mathematically manageable and conceptually elegant framework for associative memory. However, the dependence on sequential updates has long been recognized as an obstacle to efficient computation, especially in large networks. In contrast, parallel updating—in which all units are updated simultaneously—eliminates this computational bottleneck, but does so at the cost of sacrificing the ensure of monotonic descent of energy. In the classical formulation, parallel updates can induce oscillations, and to ensure convergence to a stable state, certain mathematical properties were required that limit their application in optimization problems

Several extensions of the classical Hopfield model have been proposed to overcome limitations related to storage capacity and robustness. Continuous-time formulations such as the Hopfield–Tank network \cite{Hopfield85} introduced analog neurons governed by differential equations to solve optimization problems. Other approaches have modified the learning rule to increase capacity, such as the pseudoinverse \cite{Kanter87} and the Storkey learning rules \cite{Storkey97}. The work of Krotov and Hopfield \cite{KrotovHopfield2016} introduced a generalized energy-based associative memory that extends the classical Hopfield model to achieve exponentially larger storage capacity and improved pattern separation. Their key innovation lies in redefining the energy function to include non-quadratic interactions between neurons. Building upon this generalized energy function, Demircigil et al. \cite{Demircigil_2017} extended the theory of Dense Associative Memories (DAMs) and demonstrated their robustness to noise and adversarial perturbations. More recent advances, known as modern Hopfield networks \cite{krotovHopfield2020}, reinterpret the model under the paradigm of dense associative memory and energy-based attention mechanisms, dramatically increasing representational capacity and linking Hopfield dynamics to contemporary deep learning architectures. Krotov and Hopfield \cite{krotov2021hierarchical} proposed a hierarchical extension of modern Hopfield networks aimed at enabling continuous learning and mitigating catastrophic forgetting. The work by Ramsauer et al  \cite{ramsauer2021} reformulates modern Hopfield networks within the framework of attention-based deep learning architectures, showing that the attention mechanism widely used in transformer models can be interpreted as a continuous-time Hopfield update operating in a high-dimensional space.
Although the studies cited above have succeeded in redefining the landscape of energy function or computational dynamics, resulting in improved memory capacity, the parallel computational characteristics of the network have not been thoroughly addressed.

The work of Takefuji et al. \cite{Funabiki1992NearMaxClique}  \cite{Funabiki1992} stands out, marking a pioneering milestone by proposing the first parallel Hopfield network based on the McCulloch-Pitts binary neural model to solve the Maximum Clique Problems in a reasonable computation time. This network is composed of groups of neurons and a competitive winner-take-all rule is imposed for updating the neurons. However, criticisms by Tateishi and Tamura \cite{Tateishi1994} and Wang \cite{Wang1997}  revealed limitations in the convergence of this neuronal model, which did not guarantee a decrease in the energy function. 
Building on previous approaches to parallel Hopfield dynamics, Muñoz‑Pérez and Ruiz‑Sepúlveda \cite{Munoz2011Parallelism} proposed a discrete competitive model that formalized parallel updates through neuron grouping. Their Parallelism in Binary Hopfield Networks study demonstrated that the chromatic structure of the network graph can be exploited to identify sets of neurons that can be updated simultaneously without violating energy monotonicity. This contribution provided a theoretical foundation for efficient parallel implementations and influenced subsequent developments such as the OCHOM model and other energy‑based optimization frameworks. OCHOM (Optimal Competitive HOpfield Model)  introduces a discrete Hopfield dynamics using parallel updates by groups and was developed in the early 2000s by Galán-Marín and Muñoz-Pérez \cite{GalanMarin2001OCHOM}. This model guarantees and maximizes the descent of any Lyapunov energy function, ensuring convergence to a local/global minimum while solving combinatorial problems like maximum clique. Simultaneously, Wang \cite{Wang2003} employed a parallel gradient ascent mechanism to escape local minima, while Rodés et al. \cite{Rodes2000}  developed DNNA (Discrete Neural Network Algorithm) and CNNA (Continuous Neural Network Algorithm), which are extensions of the parallel competitive Hopfield model. These algorithms prioritize guaranteed convergence over speed and are specifically designed for combinatorial optimization.

In this context, the present work explicitly focuses on the design and analysis of a parallel dynamics for Hopfield networks. Rather than modifying the underlying energy function or memory representation, we address the computational bottleneck imposed by sequential updates. By proposing and empirically validating a parallel alternative, this study contributes a complementary perspective to the existing literature and enhances the practical applicability of Hopfield-type models in modern computational settings. This model can be considered  a critical bridge to future research that integrates these dynamics into deep networks for emerging combinatorial problems.

\section{Methodology}
In Section 3 we present the mathematical developments and is divided into 5 subsections:
\begin{itemize}
    \item In 3.1, we review the concept of asynchronous dynamics, the original dynamics proposed for HNNs in 1982. \cite{Hopfield82}
    \item In 3.2, we review the concept of energy function of the network, which is a fundamental concept to work and understand the evolution of the network in time.
    \item In 3.3, we develop \cref{thm02} to understand how the value of the energy function varies when applying synchronous update to a certain set of neurons. This theorem is the basis of the new proposal in the following subsections.
    \item In 3.4, we propose a new \textbf{Discrete Differential Filter (DDF)} to facilitate synchronous operations on HNNs.
    \item In 3.5, we propose \textbf{SD-DDF} a new \textbf{Synchronous Dynamics based upon DDF}.
\end{itemize}
In Section 4 we present four computational experiments to validate the processing time enhance of the new \textbf{SD-DDF} with respect to the asynchronous dynamics.\\
\\
Finally, in Section 5 we present discussion and conclusions as well as further suggested works.\\
\\
A Hopfield Neural Network has only one layer where all neurons are connected among them in such a way that the value of each neuron affects all the others. So, the output of one neuron becomes the input for all the other neurons and, at each time instant, the exit of the whole neurons becomes the input of the whole network. This is why these neural networks are called "recurrent". The values of the neurons are changing at each time instant until the network reaches a stable state. The values of the neurons at that stable state represent the solution of the network.
The strength of the link between neuron $i$ and neuron $j$ is represented by the value $w_{ij} \in \mathbb{R}$ and is called the synaptic weight. The state of each neuron depends on the type of network to consider: Binary networks $(0: \ off;1: \ on)$, bipolar networks $(-1: \ off ; +1: \ on)$, multi-valued networks (a value of a certain set), continuous networks (value in the interval $[0,1]$), etc. Hopfield networks can evolve in time according to a discrete or a continuous time basis. Finally, if only one neuron updates its state at each time step, the network's evolution is referred to as asynchronous (or sequential) dynamics. Conversely, if multiple neurons update their states simultaneously, it is known as synchronous (parallel or partially parallel) dynamics.\\
\begin{table}[htbp]
\centering
\small
\renewcommand{\arraystretch}{1.3}
% Ajusta la tabla al ancho exacto del texto del documento
\resizebox{0.80\textwidth}{!}{%
\begin{tabular}{|c|c|c|c|c|c|c|c!{\vrule width 1.5pt}c|c|c|c|c|c|c|c|}
\hline
\multicolumn{8}{|c!{\vrule width 1.5pt}}{\textbf{Asynchronous Dynamics}} & \multicolumn{8}{c|}{\textbf{Synchronous Dynamics}} \\ \hline
\textbf{S} & $\mathbf{s_1}$ & $\mathbf{s_2}$ & $\mathbf{s_3}$ & $\mathbf{s_4}$ & $\dots$ & $\mathbf{s_n}$ & \textbf{Energy} & \textbf{S} & $\mathbf{s_1}$ & $\mathbf{s_2}$ & $\mathbf{s_3}$ & $\mathbf{s_4}$ & $\dots$ & $\mathbf{s_n}$ & \textbf{Energy} \\ \hline
$\dots$ & $\dots$ & $\dots$ & $\dots$ & $\dots$ & $\dots$ & $\dots$ & $\dots$ & $\dots$ & $\dots$ & $\dots$ & $\dots$ & $\dots$ & $\dots$ & $\dots$ & $\dots$ \\ \hline
$\mathbf{S}(t)$ & $+1$ & $+1$ & \cellcolor{highlightgreen}$+1$ & $-1$ & $\dots$ & $-1$ & $E(t)$ & $\mathbf{S}(t)$ & \cellcolor{highlightgreen}$+1$ & \cellcolor{highlightgreen}$+1$ & $+1$ & $-1$ & $\dots$ & \cellcolor{highlightgreen}$-1$ & $E(t)$ \\ \hline
$\mathbf{S}(t+1)$ & \cellcolor{highlightgreen}$-1$ & $+1$ & $+1$ & $-1$ & $\dots$ & $-1$ & $E(t+1)$ & $\mathbf{S}(t+1)$ & \cellcolor{highlightgreen}$-1$ & $+1$ & \cellcolor{highlightgreen}$-1$ & \cellcolor{highlightgreen}$+1$ & $\dots$ & \cellcolor{highlightgreen}$+1$ & $E(t+1)$ \\ \hline
$\mathbf{S}(t+2)$ & $-1$ & $+1$ & $+1$ & \cellcolor{highlightgreen}$+1$ & $\dots$ & $-1$ & $E(t+2)$ & $\mathbf{S}(t+2)$ & $-1$ & \cellcolor{highlightgreen}$-1$ & $-1$ & $+1$ & $\dots$ & \cellcolor{highlightgreen}$-1$ & $E(t+2)$ \\ \hline
$\dots$ & $\dots$ & $\dots$ & $\dots$ & $\dots$ & $\dots$ & $\dots$ & $\dots$ & $\dots$ & $\dots$ & $\dots$ & $\dots$ & $\dots$ & $\dots$ & $\dots$ & $\dots$ \\ \hline
\multicolumn{8}{|c!{\vrule width 1.5pt}}{\textbf{Updates one neuron per time step}} & \multicolumn{8}{c|}{\textbf{Updates one or more neurons per time step}} \\ \hline
\multicolumn{8}{|c!{\vrule width 1.5pt}}{\textbf{It is guaranteed that energy never increases: $E(t+1) \le E(t)$}} & \multicolumn{8}{c|}{\textbf{Objective: guarantee that the energy never increases}} \\ \hline
\end{tabular}
}
\end{table}
\\
In what follows, unless otherwise stated, we consider an HNN with bipolar-valued neurons $\{-1, +1\}$, discrete-time evolution, symmetric synaptic weights $(w_{ij}=w_{ji})$ and zero value self connections $(w_{ii}=0)$.
\subsection{Asynchronous (or Sequential) Dynamics}
Let $R(W,\theta)$ be an HNN with $n$ neurons, defined by: 
\begin{itemize}
    \item $\boldsymbol{W} \in {\mathbb{R}}^{n \times n}$ is the synaptic weight matrix, which is assumed to be symmetric and zero-diagonal: 
    \begin{equation*}
    w_{ij} = w_{ji} \quad \text{and} \quad w_{ii} = 0, \quad \forall i, j \in \{1, \dots, n\}
\end{equation*}
    \item $\boldsymbol{\theta} = [\theta_1, \dots, \theta_n]^\top$ is the threshold vector, where each $\theta_i$ represents the threshold value of the $i$-th neuron.
    \item $\mathbf{S}(t) = (s_1(t), s_2(t), \ldots, s_n(t))^ {\top}$   denote the state vector of the network at time $t$, where  $s_k(t)\in \{-1, +1\}$ represents the state of neuron \(k\).
\end{itemize}
At each time $t$ a single neuron $k$ is selected, usually at random, and its new state $s_k(t+1)$ is defined as follows: 
\[s_k(t+1)=
\begin{cases}
+1 & \text{if  } H_k(t)\geq 0 \\
-1 & \text{if  } H_k(t) < 0
\end{cases}
\]
where: \[H_k(t)=\sum_{j=1}^{n}{w_{kj}s_j(t)}-\theta_k\]
Given that $s_k(t+1)=s_k(t)+\triangle s_k(t)$, table 1 shows the values associated with the neuron $k$ during the transition from the time step $t$ to $t+1$:
\begin{table}[H]
\centering
\small
\renewcommand{\arraystretch}{1.0}
\resizebox{0.80\textwidth}{!}{%
\begin{tabular}{|c|c|c|c|c|c|}
\hline
$s_k(t+1)$ & $s_k(t)$ & $\triangle s_k(t)$ & $H_k(t)$ & $\triangle s_k(t) \cdot H_k(t)$ & $s_k(t) \cdot H_k(t) $ \\
\hline
$+1$   & $+1$   & $0$ & $\geq 0$ & $0$ & $\geq 0$ \\
$+1$  & $-1$  & $+2$ & $\geq 0$ & $\geq 0$ & $\leq 0$ \\
$-1$  & $-1$   & $0$ & $< 0$ & $0$ & $ > 0$  \\
$-1$  &  $+1$  &  $-2$ & $< 0$ & $> 0$ & $< 0$ \\
\hline
\end{tabular}
}
\caption{Change of values associated to neuron $k$ from time $t$ to $t+1$}
\end{table}
The HNN evolves over time until no neuron $k$ can change its state.  Equivalently, until: $s_k(t) H_k(t) \geq 0, \; \forall k=1,2,...,n$. 
\begin{definition}
\textbf{Asynchronous (or Sequential) Dynamics:}\\
\\
At each discrete time step $t$, a single neuron $k \in \{1,\dots,n\}$ is selected (usually at random) and updated according to its local field $H_k(t)$. The update rule is:
\[
s_k(t+1) = \operatorname{sgn}\!\big(H_k(t)\big),
\]
while the states of all other neurons remain unchanged, i.e., $s_i(t+1) = s_i(t)$ for all $i \neq k$. The dynamics proceeds asynchronously until a stable configuration is reached in which no neuron can further change its state. This condition is equivalently expressed as:
\[
s_k(t)\, H_k(t) \ge 0, \quad \forall\, k = 1,\ldots,n.
\]
\end{definition}

\subsection{Energy function}
In the current state of the network at time $t$, given by the state vector $\mathbf{S}(t) = (s_1(t), s_2(t), \ldots, s_n(t))^ {\top}$, an energy function of Lyapunov type is defined as follows:
\[E\left(t\right)=-\frac{1}{2}\sum_{i=1}^{n}\sum_{j=1}^{n}{w_{ij}s_i\left(t\right)s_j\left(t\right)}+\sum_{i=1}^{n}{\theta_is_i\left(t\right)}\]
\begin{theorem}\label{thm01}
On an HNN, the asynchronous dynamics makes the energy function decrease or stay equal at each time step. 
\begin{proof}
\begin{align}
\triangle E(t)&=E(t+1)-E(t)= \nonumber \\
&=-\frac{1}{2}\sum_{i=1}^{n}\sum_{j=1}^{n}{w_{ij}s_i(t+1)s_j(t+1)}+\sum_{i=1}^{n}{\theta_is_i(t+1)}+ \nonumber \\
&+\frac{1}{2}\sum_{i=1}^{n}\sum_{j=1}^{n}{w_{ij}s_i(t)s_j(t)}-\sum_{i=1}^{n}{\theta_is_i(t)}
\end{align}
Given that \(s_i(t+1)=s_i(t)+\triangle s_i(t)\):\\
\begin{align}
\triangle E(t)&=-\frac{1}{2}\sum_{i=1}^{n}\sum_{j=1}^{n}w_{ij}(s_i(t)+ \triangle s_i(t))(s_j(t)+ \triangle s_j(t))+ \nonumber \\
&+\sum_{i=1}^{n}\theta_i(s_i(t)+\triangle s_i(t))+\frac{1}{2}\sum_{i=1}^{n}\sum_{j=1}^{n}{w_{ij}s_i(t)s_j(t)}-\sum_{i=1}^{n}{\theta_is_i(t)}= \nonumber \\
&=-\frac{1}{2}\sum_{i=1}^{n}\sum_{j=1}^{n}w_{ij}(s_i(t) \triangle s_j(t)+s_j(t) \triangle s_i(t)+\triangle s_i(t)\triangle s_j(t))+ \nonumber \\
&+\sum_{i=1}^{n}\theta_i \triangle s_i(t)
\end{align}\\
As previously assumed, $w_{ii}=0\;\forall\,i=1,2,...n\; and \;w_{ij}=w_{ji}\;\forall i,j=1,2,...n.$ Moreover, since we are considering asynchronous dynamics, only a single unit $k$ is updated, that is,\,$ \triangle s_k(t)\neq 0\, \text{and} \, \triangle s_i(t)= 0 ;\forall i\neq k $, therefore:
\begin{align}
\triangle E(t)&=-\triangle s_k(t)(\sum_{j=1}^{n}w_{kj}s_j(t)-\theta_k)=\\
&=-\triangle s_k(t)\cdot H_k(t)\leq 0
\end{align}
by definition of asynchronous dynamics. \textbf{QED}
\end{proof}
\end{theorem}
\begin{corollary}\label{cor01}
A discrete bipolar HNN with asynchronous dynamics converges to a stable state that corresponds to a minimum, usually local, of the energy function.
\begin{proof}
    Asynchronous dynamics is designed so that each update of neurons never increases the energy of the network. Since the network has a finite number of states $(2^n)$, its energy cannot decrease forever. Therefore, the system will eventually reach a stable state in which no single neuron update can further reduce the energy. \textbf{QED}.
\end{proof}
\end{corollary}
This is a fundamental result as it guarantees the convergence of the network to a stable state thereby ensuring a solution is reached. Otherwise, the network would fail to produce a definitive output. 
\subsection{Synchronous (or Parallel or Partial Parallel) Dynamics}
\begin{theorem}\label{thm02}
    Let $R(W,\theta)$ be an HNN with bipolar-valued neurons, discrete time evolution, and symmetric and zero diagonal synaptic weight matrix $W$. Let $C$ be any subset of neurons of $R$ that would individually change their state at time $t$ according to asynchronous dynamics. By simultaneously (synchronously) updating the state of all neurons in C, the network's energy evolves according to the following expression:
    \[
    \triangle E(t)=-\sum_{i\in C}\triangle s_i(t)[\sum_{j\notin C}w_{ij}s_j(t)-\theta_i]
    \]
\begin{proof}
    In equation $(2)$, we had:
    \begin{align}
        \triangle E(t)
        &=-\frac{1}{2}\sum_{i=1}^{n}\sum_{j=1}^{n}w_{ij}(s_i(t) \triangle s_j(t)+s_j(t) \triangle s_i(t)+\triangle s_i(t)\triangle s_j(t))+ \nonumber \\
        &+\sum_{i=1}^{n}\theta_i \triangle s_i(t) \nonumber
    \end{align}
    Given the symmetry of the synaptic weights, we have
    \begin{align}
        \triangle E(t)=-\frac{1}{2}\sum_{i=1}^{n}\sum_{j=1}^{n}w_{ij}(2s_j(t)\triangle s_i(t)+\triangle s_i(t)\triangle s_j(t))+\sum_{i=1}^{n}\theta_i\triangle s_i(t)
    \end{align}
    and, after some algebra, we obtain:
    \begin{align}
        \triangle E(t)=-\sum_{i=1}^{n}\triangle s_i(t)[\sum_{j=1}^{n}w_{ij}(s_j(t)+\frac{1}{2}\triangle s_j(t))-\theta_i]
    \end{align}
    From Table 1 it is clear that \(\forall k\in C \rightarrow\; \frac{1}{2}\triangle s_k(t)=-s_k(t)\) and \(\forall k\notin C \rightarrow\; \triangle s_k(t)=0\). Therefore, the expression inside the brackets in (6) is:
    \begin{align}
        &\sum_{j=1}^{n}{w_{ij}\left(s_j\left(t\right)+\frac{1}{2}\mathrm{\Delta}s_j\left(t\right)\right)-\theta_i}=\sum_{j=1}^{n}{w_{ij}s_j\left(t\right)}-\sum_{j\in C}{w_{ij}s_j\left(t\right)}-\theta_i= \nonumber \\
        &=\sum_{j\notin C}{w_{ij}s_j\left(t\right)}-\theta_i\ \ \ \ 
    \end{align}
    And, substituting (7) in (6), we have that:
    \begin{align}
        \triangle E(t)=-\sum_{i\in C}\triangle s_i(t)[\sum_{j\notin C}w_{ij}s_j(t)-\theta_i].\textbf{QED}
    \end{align}
\end{proof}
\end{theorem}
\cref{thm02} shows that the contribution of each neuron selected for updating at time $t$ to the change in energy of the network is equal to the increment in the neuron's value multiplied by the contribution received from the neurons not selected for updating. This property is already implicit in asynchronous dynamics and \cref{thm02} makes it explicit.
\begin{corollary}\label{cor02}
    Expression (8) represents the natural generalization of the network's energy change  at time $t$, which extends it from asynchronous (one neuron) to synchronous (more than one neuron) dynamics.
    \begin{proof}
        In asynchronous dynamics, only one neuron $k$ is selected for change at time $t$. In that case \(C=\{k\}\) and, therefore:
        \begin{align}
            \triangle E(t)&=-\sum_{i\in C}\triangle s_i(t)[\sum_{j\notin C}w_{ij}s_j(t)-\theta_i]= \nonumber \\
            &= -\triangle s_k(t)[\sum_{j \neq k}w_{ij}s_j(t)-\theta_i]= -\triangle s_k(t)H_k(t) \leq 0 
        \end{align}
    like in (4), by the definition of asynchronous dynamics. \textbf{QED}
    \end{proof}
\end{corollary}
In \cref{thm01}, it was demonstrated that, under asynchronous dynamics, the energy of the network never increases. However, for synchronous dynamics, \cref{thm02} does not allow us to make the same claim for any given set of neurons C.
\begin{corollary}\label{cor03}
For any given set $C$ of neurons in $R$ that would individually change their state at time $t$ according to asynchronous dynamics, it cannot be guaranteed that, by synchronously updating  all neurons in $C$, the energy of the network will decrease or remain the same. 
\begin{proof}
From \cref{thm02}:
\begin{align}
    \triangle E(t)&=-\sum_{i\in C}\triangle s_i(t)[\sum_{j\notin C}w_{ij}s_j(t)-\theta_i]= \nonumber \\
    &=-\sum_{i\in C}\triangle s_i(t)[\sum_{j\notin C}w_{ij}s_j(t)-\theta_i+\sum_{j\in C}w_{ij} s_j(t)-\sum_{j\in C}w_{ij} s_j(t)]= \nonumber \\
    &=-\sum_{i\in C}\triangle s_i(t)[\sum_{j=1}^{n}w_{ij}s_j(t)-\theta_i]+\sum_{i\in C}\sum_{j\in C}w_{ij}\triangle s_i(t)s_j(t)=\nonumber \\
    &=-\sum_{i \in C}\triangle s_i(t)H_i(t)+\sum_{i\in C}\sum_{j\in C}w_{ij}\triangle s_i(t)s_j(t)
\end{align}
Because all neurons in $C$ satisfy the asynchronous dynamics criterion, the first term in $(10)$ will be less than or equal to zero, but, in general, one cannot guaranty that the whole expression (10) is less than or equal to zero. \textbf{QED}
\end{proof}
\end{corollary}
In summary, we have a set $C$ of neurons such that, updating the state of any single neuron guarantees that the network's energy will not increase. However, if these neurons are updated synchronously, we can no longer guarantee that the energy will decrease or remain constant. To overcome this situation, we introduce a novel tool for HNNs in the next section.
\subsection{Proposal of a new Discrete Differential Filter (DDF)}
Given any set $C$ of neurons in $R(W,\theta)$, which would individually change their state at time $t$ according to asynchronous dynamics, our goal is to select a subset $C'\subset C$ that guaranties that, by changing synchronously in $R$ the state of all neurons in $C'$, the energy of the network decreases as much as possible or remains unchanged. \\
\\
All possible subsets $C'$ belong to the power set $\mathcal{P}(C)$ so, the number of candidate subsets $C'$ is $2^{|C|}$. Determining a suitable $C'$ is itself a  combinatorial optimization problem. To select $C'$, we introduce the concept of \textbf{Discrete Differential Filter (DDF)} of $R(W, \theta)$ at $\mathbf{S}(t)$ for $C$:\\
\\
Expression (10) can be rewritten as
\begin{align}
    \triangle E(t)&=-\sum_{i \in C}\triangle s_i(t)H_i(t)+\sum_{i\in C}\sum_{j\in C}w_{ij}\triangle s_i(t)s_j(t)= \nonumber \\
    &=2\sum_{i\in C} s_i\left(t\right)H_i\left(t\right)-2\sum_{i\in C}\sum_{j\in C}{w_{ij}s_i\left(t\right)s_j\left(t\right)}= \nonumber \\
    &=-\frac{1}{2}\sum_{i\in C}\sum_{j\in C}{4w_{ij}}s_i\left(t\right)s_j\left(t\right)+\sum_{i\in C}{2H_i}\left(t\right)s_i\left(t\right)=E^\prime
\end{align}
Expression (11) can be interpreted as the energy function of a new Hopfield network $R'(W', \theta')$.
\begin{definition}{\textbf{Discrete Differential Filter (DDF)}}\\
    The Hopfield network $R'(W', \theta')$ with $m$ binary-valued neurons $\{0, 1\}$, with state vector $X$, defined from $R(W, \theta)$ and $\mathbf{S}(t)$ for $C$ as follows:
\begin{align}
    &w'_{ij}=4w_{ij}s_i(t)s_j(t), \ \forall i,j \in C\\
    &\theta'_i=2H_i(t)s_i(t), \ \forall i \in C \\
    &m=|C| \\
    &X(t')=(x_i(t'))\in \{0,+1\}^m
\end{align}
is called the \textbf{Discrete Differential Filter (DDF)} of $R(W, \theta)$ at $\mathbf{S}(t)$ for $C$ which will be denoted as DDF(R,$\mathbf{S}(t)$,$C$).
\end{definition}
Note that (13) is the change of energy of the bipolar network $R$ if "$i$" is the only one neuron that changes its state at time $t$. In effect, in (4) it was stated that when only one neuron changes its state, then $\Delta E(t)=-\Delta s_k(t)·H_k(t)$ and $\frac{1}{2}\Delta s_k(t)=-s_k(t)$.\\
\\
The new matrix of synaptic weights $W'$ is symmetric $(w'_{ij}=w'_{ji})$ because $(w_{ij}s_i(t)s_j(t)=w_{ji}s_j(t)s_i(t))$ and  zero-diagonal $(w'_{ii}=0)$ because ($w_{ii}=0$). The energy function of $R'$ is given as follows:
\begin{align}
    E^\prime=-\frac{1}{2}\sum_{i=1}^{m}\sum_{j=1}^{m}{w^\prime}_{ij}x_ix_j+\sum_{i=1}^{m}{{\theta^\prime}_ix_i}
\end{align}
The \textbf{DDF} network $R'$ is a binary Hopfield network. The solution $\mathbf{X}=(x_r)_{r=1}^m$ of $R'$ yields the subset $C'=\{r\in C \mid x_r=1\}\subset C$, represents the combination of neurons of $C$ that produces an optimal decrease in the energy of $R$ when their states change synchronously. \\
\\
$R'$ will be solved using asynchronous dynamics; however, rather than selecting a random initial state, \textbf{DDF} will employ a progressive neuron activation scheme starting from the zero state $\mathbf{X}(0)=(x_r(0)=0)_{r=1}^m$. This is done by applying the following variant of the asynchronous update rule: at each time step $t'$, a neuron $r \in C$ satisfying $x_r(t')=0$ is randomly selected, and the following activation criteria will be applied:\\
\[x_r(t'+1)=
\begin{cases}
+1 & \text{if  } H'_r(t')\geq 0 \ \And \ x_r(t')=0\\
0 & \text{if  } H'_r(t') < 0
\end{cases}
\]
where:
\begin{align}
    H'_r(t')=\sum_{j=1}^{m}w'_{rj}x_j(t')-\theta'_r
\end{align}
At each time step $t'$ the energy $E'$ will decrease or remain the same as the update follows standard asynchronous dynamics except for the criteria to choose neurons with $x_r(t')=0$. This process terminates when no neurons remains in $C$ satisfy the aforementioned selection criteria. Although this final state is not guaranteed to be a local minimum, it will serve as the initial state for the subsequent application of standard asynchronous dynamics. From this state, asynchronous dynamics is applied until a local minimum is reached; specifically: at time $t'$, a single neuron $x_r$ is chosen at random and its state is updated according to:
\[x_r(t'+1)=
\begin{cases}
+1 & \text{if  } H'_r(t')\geq 0 \\
0 & \text{if  } H'_r(t') < 0
\end{cases}
\]
where:
\begin{align}
    H'_r(t')=\sum_{j=1}^{m}w'_{rj}x_j(t')-\theta'_r
\end{align}
\begin{lemma}\label{lem01}
Let $C$ be any set of neurons of $R$ that would change their states at time $t$ according to asynchronous dynamics and let the zero vector $(x_k(0))_{k=1}^m$ with $x_k(0)=0$ for all $k$ be the initial state of $R'=DDF(R,\mathbf{S},C)$. Then, under asynchronous dynamics in $R'$, any neuron $r \in C$ chosen at time step $t'=1$ will be activated, that is,  $x_r(1)=+1$.\\
\\
\textbf{Proof:} Indeed, let $x_k(0)=0,\ \forall k=1,2,...,m$ and let $r$ be the first neuron of $C$ chosen at random at time $t'=1$. According to asynchronous dynamics in $R'$, the value of $x_r(1)$ depends on the value of $H'_r(0)$, which is:
\[
    H'_r(0)=\sum_{j=1}^{m}w'_{rj}x_j(0)-\theta'_r
    =-\theta'_r=-2H_r(t)s_r(t)\geq 0
\]
And, therefore: $x_r(1) = \operatorname{sgn}\!\big(H'_r(0)\big)=+1$. \textbf{QED}
\end{lemma}
This result ensures that any non-empty set $C$ yields to a non-empty set $C'$ containing at least one neuron. Moreover, if $C$ contains only one neuron, then $C'=C$ and its state  change causes the energy in $R$ to decrease or remain unchanged.
\begin{theorem}\label{thm03}
    Upon reaching a solution (local minimum) $\mathbf{X}=(x_i)\in \{0,+1\}^m$ for $R'=DDF(R,\mathbf{S},C)$, neurons for which $x_i=+1$ are selected from $C$ to form the subset $C'$. Changing synchronously in $R$ at time $t$ the state of all neurons of $C'$, the following holds:
    \begin{itemize}
        \item the change in energy of $R$ is equal to the value of $E'$ reached at state $\mathbf{X}$ and
        \item the change in energy of $R$ is a maximum decrease at time $t$ or the energy will remain the same.
    \end{itemize}
    \begin{proof}
             In effect, as has been stated in $(11)$:
            \begin{align}
            E'&=-\frac{1}{2}\sum_{i=1}^{m}\sum_{j=1}^{m}{w'}_{ij}x_ix_j+\sum_{i=1}^{m}{{\theta^\prime}_ix_i}= \nonumber \\
            &=-\frac{1}{2}\sum_{i \in C'}\sum_{j \in C'}w'_{ij}x_ix_j+\sum_{i \in C'} \theta'_ix_i= \nonumber \\
            &=-\frac{1}{2}\sum_{i \in C'}\sum_{j \in C'}4w_{ij}s_i(t)s_j(t) + \sum_{i \in C'}2H_i(t)s_i(t)=\triangle E(t) \nonumber
        \end{align}
         which proves the first statement.\\
\\
 For the second statement, let $\mathbf{X}$  be the local minimum reached in $R'$. \cref{lem01} shows that  $\mathbf{X}$ contains at least one neuron with a value +1. There are two possible cases:\\
        \\
        Case I: If $\exists! r \mid x_r=+1 \; and \;x_q =0\; \forall q\neq r$. Then:
        \begin{align}
            \triangle E = E'&=-\frac{1}{2}\sum_{i=1}^{m}\sum_{j=1}^{m}{w^\prime}_{ij}x_ix_j+\sum_{i=1}^{m}{{\theta^\prime}_ix_i}=\theta'_r = \nonumber \\
            &= 2H_r(t)s_r(t) = -\triangle s_r(t)H_r(t) \leq 0 \nonumber
        \end{align}
        given that $r \in C$ and satisfies the asynchronous dynamics criterion in $R$.\\
        \\
        Case II: If exists more than one neuron in $\mathbf{X}$ with value $+1$. Let $C'$ be the set of neurons with value $+1$ in $\mathbf{X}$. Since only one neuron with value $+1$ makes $E' \leq 0$ and in each iteration in $R'$, $E'$ will decrease or remain the same, in the final value of $\mathbf{X}$, $E'$ will be $\leq 0$ . Then:
        \begin{align}
            \triangle E = E'\leq 0 \nonumber
        \end{align}
As far as the value of $E'$ corresponds to a minimum in $R'$, the change of energy of $R$ will be a maximum decrease or will remain the same.\textbf{QED}
    \end{proof}
\end{theorem}
\begin{corollary}\label{cor04}
    From $t'=1$, and at all subsequent times, the value of $E'$ in $R'=DDF(R,\mathbf{S},C)$ is less than or equal to zero. Therefore, the decrease in the energy function in $R$ when synchronously changing the value of all neurons in $C'$ at any time $t' \geq 1$ will be less than or equal to zero, even though $\mathbf{X}$ is not a local minimum.
    \begin{proof}
        This result follows directly from \cref{thm03}.
    \end{proof}
\end{corollary}
This corollary allows to stop the process of obtaining a set $C' \subset C$ that always makes the energy function in $R$ to decrease, even though $\mathbf{X}$ is not a local minimum of $R'$. In this case, the decrease of energy in $R$ will not be maximal but it will still decrease. So that, the number of time steps (instants) to obtain $\mathbf{X}$ can be limited in order to optimize the trade-off between speed and results of each $R'$ and those of the overall $R$.\\
\\
\textbf{DDF} can be used for different purposes because is a tool to help to pick up a set of neurons that, by changing their states synchronously at a time step t, the energy of the network decreases or remain the same.
\subsection{SD-DDF: New Synchronous Dynamics based upon DDF}
In this paper we shall use it to introduce a new synchronous dynamics based upon \textbf{DDF}.\\
\\
Let $R(W,\theta)$ be an HNN, discrete, bipolar, symmetric $W$ and zero value self connections. Based upon \textbf{DDF}, we define the following dynamics:
\begin{itemize}
    \item Step 1: At time $t$ and state $\mathbf{S}(t)=(s_k(t))$, select $C$ as the set of \textbf{all} neurons of $R$ that would change their state according to the condition of asynchronous dynamics: $C=\{k\in R \mid s_k(t)·H_k(t)\leq0\ \forall k=1,2,...,n\}$. The process finishes when $C=\emptyset$, in other words, when there are no neurons in $R$ eligible for an update.
    \item Step 2: Construct $R'=DDF(R,\mathbf{S}(t),C)$ according to (12), (13), (14) and (15).
    \item Step 3: Obtain $\mathbf{X}$ as the solution to $R'$ and get $C'\subset C$ with the neurons of $C$ corresponding to values $+1$ in $\mathbf{X}$. This process may take until $\mathbf{X}$ becomes a local minimum of $R'$ or, according to \cref{cor04}, until the limit of time steps (time) is reached.
    \item Step 4: Update synchronously in $R$ all neurons in $C'$ and go to Step 1.
\end{itemize}
Based on previously proved theorems and corollaries, this sequence of steps is guaranteed to converge to a local minimum. Moreover, it performs this convergence faster than asynchronous dynamics: at each time step $t$, it performs an optimal jump within the network $R$, since $C$ contains all neurons eligible to change according to the asynchronous dynamics criterion, and $C'\subset C$ is selected by \textbf{DDF} to optimize the decrease of the energy function. If the process is stopped at step 3 before reaching a minimum, according to \cref{cor04}, the energy of the network simply will decrease or stay the same.\\
\\
\textbf{DDF} shows one way to solve Hopfield networks by using Hopfield networks. At each time step, \textbf{SD-DDF} explores the network and selects the optimal set of neurons to change synchronously to produce an optimal energy descent in $R$ or let the energy remain the same. Exploring the network part by part does not require any preconditions regarding its structure or dynamics organization, as \textbf{SD-DDF} autonomously selects the subset of the network to explore, namely, the set $C$.\\
In the next section, we evaluate the newly proposed synchronous dynamics \textbf{SD-DDF} against asynchronous dynamics.

\section{Computational Experiments}
The aim of this section is to evaluate the ability of the proposed SD-DDF dynamics to find local minima in processing time shorter than that of the asynchronous dynamics while obtaining similar values for the rest of variables considered like energy, etc. Four tests have been performed, each addressing one problem: Graph Bi-partition, Random Generated Network, N-Queen allocation and Travelling Salesman Problem (TSP).\\
For each problem and for each number of neurons, five instances of the problem were generated and each instance was solved thirty times per each dynamics. All runs were carried out until local minimums were reached, so the standard deviation values do not reflect the measurement error, but rather the   dispersion of actual local minima obtained. Key variables were collected in each run. In all cases, symmetric synaptic weight matrices and zero diagonal were considered.\\
The experiments were performed on Jupiter notebooks running on the anaconda.com online platform. The processing times reported are not intended to be compared with external results, but only among the dynamics evaluated in this study.
Three dynamics were considered in these experiments:
\begin{enumerate}
    \item SEQ. The fundamental asynchronous dynamics of Hopfield: at each time $t$ one neuron $k$ is selected randomly and the change is evaluated according to the value of $\mathbf{H_k(t)}$.
    \item C-SEQ. The SEQ dynamics is conceptually simple and powerful, though is getting slower as the network grows in size and / or  the computation is arriving to a minimum as the probability of find randomly one neuron which change of state makes the energy function to decrease or remain equal is gradually smaller. To compensate this effect, the C-SEQ dynamics is introduced which consists of: at each time $t$, a set $C$ of neurons that would change individually according to the SEQ dynamics is obtained, and C-SEQ randomly chooses a neuron from $C$ and directly changes its state.
    \item \textbf{SD-DDF} (Synchronous Dynamics based upon Discrete Differential Filter). The new dynamics proposed. At each time $t$, a set $C$ of neurons that would change individually according to the SEQ dynamics is obtained and $R'=DDF(R,\mathbf{S},C)$ is constructed and solved. The neurons $x_r=+1$ in the solution of $R'$ are the neurons of $R$ chosen to change their states simultaneously at time $t$. For these experiments, a limit of 100 time steps has been established (see \cref{cor04}).
\end{enumerate}
\subsection{Graph bipartition (1-BP)}
Let $G(V,E)$ be a graph without self-loops, where $V$ is the set of vertices and $E$ the set of edges, the graph bipartition problem consists of dividing $V$ into two subsets $V_1$ and $V_2$ so that the two subsets contain approximately the same number of vertices, while minimizing and the number of edges connecting the vertices belonging to different subsets is minimum.\\
To model the problem, let $\mathbf{S}= [s_1,s_2...,s_n]^T$ be the vector state with $s_i$  defined per each vertex $i$ such that:
\[
s_i =
\begin{cases}
+1, & \text{if vertex $i$ belongs to $V_1$} \\
-1, & \text{if vertex $i$ belongs to $V_2$}
\end{cases}
\]
Let $A$ be the adjacency matrix of $G$ where:
\[
a_{ij} =
\begin{cases}
    1, & \text{if exists and edge between vertex $i$ and vertex $j$}\\
    0, & \text{otherwise}
\end{cases}
\]
To minimize the number of edges between $V_1$ and $V_2$ it is necessary to minimize:
\[
\sum_{i=1}^{n} \sum_{j=1}^{n} a_{ij} \frac{1 - s_i s_j}{2}
\]
And to make equal the number of vertices of each subset, we minimize:
\[
\lambda \left( \sum_{i=1}^{n} s_i \right)^2
\]
Where $\lambda$ is the Lagrange multiplier to define the relative importance of this restriction against the other. %In this case, $\lambda = 0.5$ has been used.
So that, the function to minimize is:
\[
E= \sum_{i=1}^{n} \sum_{j=1}^{n} a_{ij} \frac{1 - s_i s_j}{2} 
    + \lambda \left( \sum_{i=1}^{n} s_i \right)^2
\]
Operating the above, we obtain:
\[
E=-\frac{1}{2}\sum_{i=1}^n\sum_{j=1, j \neq i }^n(a_{ij}-2\lambda)s_is_j+Constants
\]
and, comparing with the standard form of the energy function:
\[
E=-\frac{1}{2}\sum_{i=1}^n\sum_{j=1}^nw_{ij}s_is_j+\sum_{i=1}^n\theta_is_i
\]
it follows the definition of the coefficients of the synaptic weight matrix $W = (\omega_{ij})_{n \times n}$ and the threshold vector values $\boldsymbol{\theta} = (\theta_1, \dots, \theta_n)^T$\ \ as: \(\omega_{ij} = a_{ij} - 2\lambda,\
with \ \omega_{ii} = 0\ and \ \ \theta_i = 0,\ \forall i \).\\For the experiments designed to evaluate the three aforementioned dynamics, the Lagrange multiplier was set to $\lambda=0.5$.
A random Erdos-Renyi graph has been generated with a probability of edge equal to 0.20 and a seed equal to 42. 
To evaluate the quality of the solutions reached in each run, three variables have been used: energy, processing time and the normalized cut $NC(G)$ defined as follows: Being $G(V,E),\ \ V=V_1\cup V_2$:
\[
NC(G) = \operatorname{cut}(G) \left( \frac{1}{\operatorname{vol}(V_1)} + \frac{1}{\operatorname{vol}(V_2)} \right)
\]
\[
\operatorname{cut}(G) = \sum_{\substack{(v,w) \in E \\ v \in V_1,\ w \in V_2}} 1
\]
\[
\operatorname{vol}(V_1) = \sum_{v \in V_1} \deg(v)
\]
\[
\deg(v) = \sum_{(v,w) \in E} 1
\]

\begin{table}[H]
\centering
\tiny
\resizebox{\textwidth}{!}{%
\renewcommand{\arraystretch}{0.8}
\begin{tabular}{|l|r|r|r|r|r|}
\hline
 & \multicolumn{3}{c|}{DYNAMICS} & \multicolumn{2}{c|}{VARIATION} \\ \hline
NEURONS & SEQ & C-SEQ & SD-DDF & SD-DDF/SEQ - 1 & SD-DDF/C-SEQ - 1 \\ \hline
\textbf{100} & & & & & \\ \hline
energy (avg) & -270.51 & \textbf{-282.01} & -275.63 & 0.0189 & -0.0226 \\ \hline
energy (min) & -312.00 & -328.00 & \textbf{-332.00} & 0.0641 & 0.0122 \\ \hline
energy (desvest) & 21.90 & \textbf{18.95} & 19.24 & -0.1217 & 0.0152 \\ \hline
time (avg) & 0.1734 & 0.0264 & \textbf{0.0111} & -0.9359 & -0.5796 \\ \hline
time (min) & 0.0238 & 0.0082 & \textbf{0.0030} & -0.8748 & -0.6368 \\ \hline
time (desvest) & 0.1600 & 0.0180 & \textbf{0.0072} & -0.9548 & -0.5980 \\ \hline
NC(G) (mean) & \textbf{0.7745} & 0.7625 & 0.7691 & -0.0069 & 0.0087 \\ \hline
NC(G) (min) & \textbf{0.7316} & 0.7155 & 0.7111 & -0.0280 & -0.0062 \\ \hline
NC(G) (desvest) & 0.0226 & \textbf{0.0195} & 0.0197 & -0.1253 & 0.0131 \\ \hline
\# of valid cases & 150 & 137 & 146 &  &  \\ \hline
\textbf{200} & & & & & \\ \hline
energy (avg) & -804.83 & \textbf{-811.23} & -791.81 & -0.0162 & -0.0239 \\ \hline
energy (min) & -877.00 & \textbf{-887.00} & -863.00 & -0.0160 & -0.0271 \\ \hline
energy (desvest) & 37.30 & \textbf{32.98} & 34.59 & -0.0727 & 0.0488 \\ \hline
time (avg) & 0.6901 & 0.0714 & \textbf{0.0305} & -0.9559 & -0.5734 \\ \hline
time (min) & 0.1961 & 0.0256 & \textbf{0.0079} & -0.9599 & -0.6927 \\ \hline
time (desvest) & 0.3409 & 0.0487 & \textbf{0.0177} & -0.9480 & -0.6353 \\ \hline
NC(G) (mean) & 0.8260 & \textbf{0.8245} & 0.8293 & 0.0040 & 0.0058 \\ \hline
NC(G) (min) & 0.8082 & \textbf{0.8059} & 0.8113 & 0.0038 & 0.0068 \\ \hline
NC(G) (desvest) & 0.0092 & \textbf{0.0082} & 0.0086 & -0.0713 & 0.0502 \\ \hline
\# of valid cases & 150 & 121 & 150 &  &  \\ \hline
\textbf{400} & & & & & \\ \hline
energy (avg) & -2270.99 & \textbf{-2296.58} & -2246.01 & -0.0110 & -0.0220 \\ \hline
energy (min) & -2425.00 & \textbf{-2455.00} & -2381.00 & -0.0181 & -0.0301 \\ \hline
energy (desvest) & 76.99 & \textbf{67.63} & 78.73 & 0.0227 & 0.1641 \\ \hline
time (avg) & 3.5222 & 0.1950 & \textbf{0.0638} & -0.9819 & -0.6727 \\ \hline
time (min) & 1.0491 & 0.1012 & \textbf{0.0190} & -0.9819 & -0.8124 \\ \hline
time (desvest) & 1.6170 & 0.0781 & \textbf{0.0382} & -0.9764 & -0.5112 \\ \hline
NC(G) (mean) & 0.8716 & \textbf{0.8700} & 0.8732 & 0.0018 & 0.0036 \\ \hline
NC(G) (min) & 0.8620 & \textbf{0.8602} & 0.8648 & 0.0033 & 0.0053 \\ \hline
NC(G) (desvest) & 0.0048 & \textbf{0.0042} & 0.0049 & 0.0210 & 0.1625 \\ \hline
\# of valid cases & 150 & 119 & 150 &  &  \\ \hline
\textbf{800} & & & & & \\ \hline
energy (avg) & -6357.99 & \textbf{-6380.79} & -6268.91 & -0.0140 & -0.0175 \\ \hline
energy (min) & \textbf{-6677.00} & -6639.00 & -6635.00 & -0.0063 & -0.0006 \\ \hline
energy (desvest) & 141.59 & 144.78 & \textbf{137.72} & -0.0274 & -0.0487 \\ \hline
time (avg) & 14.7663 & 0.7678 & \textbf{0.1907} & -0.9871 & -0.7516 \\ \hline
time (min) & 6.1614 & 0.4023 & \textbf{0.0665} & -0.9892 & -0.8346 \\ \hline
time (desvest) & 4.9862 & 0.2670 & \textbf{0.1161} & -0.9767 & -0.5650 \\ \hline
NC(G) (mean) & 0.9070 & \textbf{0.9066} & 0.9084 & 0.0015 & 0.0019 \\ \hline
NC(G) (min) & \textbf{0.9020} & 0.9026 & 0.9027 & 0.0007 & 0.0001 \\ \hline
NC(G) (desvest) & 0.0022 & 0.0023 & \textbf{0.0021} & -0.0291 & -0.0496 \\ \hline
\# of valid cases & 150 & 248 & 300 &  &  \\ \hline
\textbf{1600} & & & & & \\ \hline
energy (avg) & & \textbf{-18112.53} & -17809.63 & & -0.0167 \\ \hline
energy (min) & & \textbf{-18529.00} & -18439.00 & & -0.0049 \\ \hline
energy (desvest) & & \textbf{209.29} & 324.69 & & 0.5514 \\ \hline
time (avg) & & 4.5212 & \textbf{0.7699} & & -0.8297 \\ \hline
time (min) & & 2.9984 & \textbf{0.3767} & & -0.8744 \\ \hline
time (desvest) & & 1.1184 & \textbf{0.2659} & & -0.7622 \\ \hline
NC(G) (mean) & & \textbf{0.9323} & 0.9334 & & 0.0013 \\ \hline
NC(G) (min) & & \textbf{0.9306} & 0.9310 & & 0.0004 \\ \hline
NC(G) (desvest) & & \textbf{0.0008} & 0.0013 & & 0.5516 \\ \hline
\# of valid cases & & 129 & 150 & &  \\ \hline
\end{tabular}%
}
\caption{BP experiment results. Best results per row in bold}
\end{table}
In the case of the BP problem, Table 2 shows that the advantage in processing time (avg) of \textbf{SD-DDF} with respect to SEQ is for all network sizes over 90\% getting over 98\% for 800 neurons. With respect to C-SEQ the advantage starts at 57.96\% for networks with 100 neurons and goes up to 82.97\% for networks with 1600 neurons. In general, \textbf{SD-DDF} is reaching similar results in the other variables except in those related to processing time where \textbf{SD-DDF} is much superior to SEQ and C-SEQ.\\
Figure 2 shows a representation of the time (avg) variable for the different dynamics and network sizes showing that for the network sizes considered, the trend of time (avg) grows polynomially in the three cases, but with a significant difference in the coefficient of $x^2$: $2.43\cdot 10^{-5}$ (SEQ), $2.37\cdot 10^{-6}$ (C-SEQ) and $3.08\cdot 10^{-7}$ (\textbf{SD-DDF}).\\
%------------------
\begin{figure}[H]
    \centering

\begin{tikzpicture}
\begin{semilogyaxis}[
    xlabel={Neurons},
    ylabel={Processing Time (s)},
    legend style={
        font=\footnotesize,
        cells={anchor=west},
        at={(0.5,-0.2)},
        anchor=north,
        legend columns=1
    },
    grid=both,
    grid style={line width=0.1pt, draw=gray!30},
    major grid style={line width=0.2pt, draw=gray!50},
    width=10cm,
    height=7cm,
    xmin=50, xmax=1700,
    ymin=0.005, ymax=20,
    xtick={100, 200, 400, 800, 1600},
    log ticks with fixed point
]

% SEQ - Data points
\addplot[only marks, mark=square*, mark size=3pt, blue]
    coordinates {(100,0.1734) (200,0.6901) (400,3.5222) (800,14.7663)};
\addlegendentry{SEQ}

% SEQ - Polynomial trendline
\addplot[blue, dashed, thick, domain=100:800, samples=100]
    {2.43e-05*x^2 - 0.000987*x - 0.0129};
\addlegendentry{$y = 2.43{\cdot}10^{-5}x^2 - 9.87{\cdot}10^{-4}x - 1.29{\cdot}10^{-2}$, $R^2=0.9999$}

% C-SEQ - Data points
\addplot[only marks, mark=triangle*, mark size=3pt, red]
    coordinates {(100,0.0264) (200,0.0714) (400,0.195) (800,0.7678) (1600,4.5212)};
\addlegendentry{C-SEQ}

% C-SEQ - Polynomial trendline
\addplot[red, dashed, thick, domain=100:1600, samples=100]
    {2.37e-06*x^2 - 0.00107*x + 0.166};
\addlegendentry{$y = 2.37{\cdot}10^{-6}x^2 - 1.07{\cdot}10^{-3}x + 1.66{\cdot}10^{-1}$, $R^2=0.9991$}

% SD-DDF - Data points
\addplot[only marks, mark=o, mark size=3pt, green!60!black]
    coordinates {(100,0.0111) (200,0.0305) (400,0.0638) (800,0.1907) (1600,0.7699)};
\addlegendentry{SD-DDF}

% SD-DDF - Polynomial trendline
\addplot[green!60!black, dashed, thick, domain=100:1600, samples=100]
    {3.08e-07*x^2 - 2.32e-05*x + 0.0175};
\addlegendentry{$y = 3.08{\cdot}10^{-7}x^2 - 2.32{\cdot}10^{-5}x + 1.75{\cdot}10^{-2}$, $R^2=0.9996$}

\end{semilogyaxis}
\end{tikzpicture}
\caption{BP Problem: Processing Time vs Network Size}
\label{fig:mi_grafica_tikz}
\end{figure}
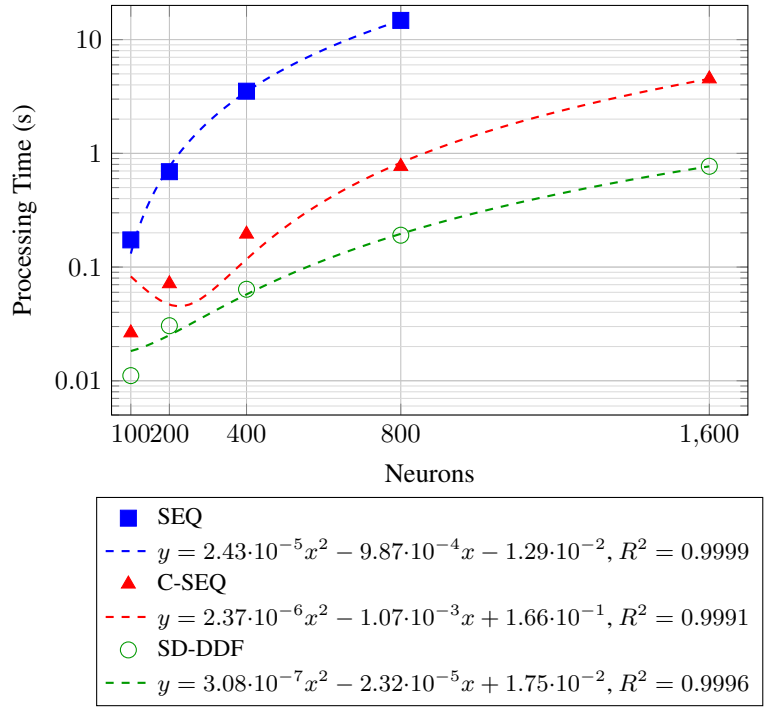
%------------------
\subsection{Random generated network (2-RG)}
A random synaptic weight matrix is generated with coefficients sampled from a standard normal distribution, $N(0, 1)$. The threshold vector is set to zero.
Energy and processing time variables have been used to evaluate the quality of the local minimum reached.
\begin{table}[H] %[htbp]
\centering
\tiny
\resizebox{\textwidth}{!}{%
\renewcommand{\arraystretch}{0.8}
\begin{tabular}{|l|r|r|r|r|r|}
\hline
 & \multicolumn{3}{c|}{DYNAMICS} & \multicolumn{2}{c|}{VARIATION} \\ \hline
NEURONS & SEQ & C-SEQ & SD-DDF & SD-DDF/SEQ - 1 & SD-DDF/C-SEQ - 1 \\ \hline
\textbf{100} & & & & & \\ \hline
energy (avg) & \textbf{-463.93} & -463.06 & -458.69 & -0.0113 & -0.0094 \\ \hline
energy (min) & \textbf{-522.61} & -509.06 & -516.60 & -0.0115 & 0.0148 \\ \hline
energy (desvest) & 25.63 & \textbf{22.73} & 26.39 & 0.0295 & 0.1610 \\ \hline
time (avg) & 0.2357 & 0.0476 & \textbf{0.0115} & -0.9513 & -0.7591 \\ \hline
time (min) & 0.0525 & 0.0078 & \textbf{0.0037} & -0.9289 & -0.5216 \\ \hline
time (desvest) & 0.1766 & 0.0575 & \textbf{0.0070} & -0.9601 & -0.8774 \\ \hline
\# of valid cases & 150 & 150 & 148 &  &  \\ \hline
\textbf{200} & & & & & \\ \hline
energy (avg) & \textbf{-1350.86} & -1344.66 & -1329.73 & -0.0156 & -0.0111 \\ \hline
energy (min) & \textbf{-1473.44} & -1465.43 & -1458.12 & -0.0104 & -0.0050 \\ \hline
energy (desvest) & \textbf{55.48} & 58.44 & 64.88 & 0.1693 & 0.1100 \\ \hline
time (avg) & 0.8975 & 0.0687 & \textbf{0.0251} & -0.9720 & -0.6346 \\ \hline
time (min) & 0.2503 & 0.0298 & \textbf{0.0106} & -0.9575 & -0.6428 \\ \hline
time (desvest) & 0.4492 & 0.0405 & \textbf{0.0174} & -0.9612 & -0.5701 \\ \hline
\# of valid cases & 150 & 150 & 149 &  &  \\ \hline
\textbf{400} & & & & & \\ \hline
energy (avg) & -3828.70 & \textbf{-3831.43} & -3783.04 & -0.0119 & -0.0126 \\ \hline
energy (min) & -4082.09 & \textbf{-4082.97} & -4024.00 & -0.0142 & -0.0144 \\ \hline
energy (desvest) & \textbf{100.09} & 113.01 & 103.02 & 0.0292 & -0.0884 \\ \hline
time (avg) & 3.5876 & 0.1661 & \textbf{0.0521} & -0.9855 & -0.6861 \\ \hline
time (min) & 1.1396 & 0.1067 & \textbf{0.0255} & -0.9776 & -0.7607 \\ \hline
time (desvest) & 1.5197 & 0.0527 & \textbf{0.0272} & -0.9821 & -0.4831 \\ \hline
\# of valid cases & 150 & 150 & 146 &  &  \\ \hline
\textbf{800} & & & & & \\ \hline
energy (avg) & & \textbf{-10953.03} & -10878.04 & & -0.0068 \\ \hline
energy (min) & & \textbf{-11602.23} & -11542.36 & & -0.0052 \\ \hline
energy (desvest) & & \textbf{235.23} & 253.45 & & 0.0775 \\ \hline
time (avg) & & 0.7315 & \textbf{0.1663} & & -0.7727 \\ \hline
time (min) & & 0.4442 & \textbf{0.0683} & & -0.8462 \\ \hline
time (desvest) & & 0.2027 & \textbf{0.0920} & & -0.5462 \\ \hline
\# of valid cases & & 150 & 150 & &  \\ \hline
\textbf{1600} & & & & & \\ \hline
energy (avg) & & \textbf{-31266.28} & -31039.28 & & -0.0073 \\ \hline
energy (min) & & \textbf{-32823.09} & -32367.88 & & -0.0139 \\ \hline
energy (desvest) & & \textbf{447.28} & 495.63 & & 0.1081 \\ \hline
time (avg) & & 4.4090 & \textbf{0.5427} & & -0.8769 \\ \hline
time (min) & & 2.9559 & \textbf{0.2651} & & -0.9103 \\ \hline
time (desvest) & & 0.9310 & \textbf{0.2016} & & -0.7835 \\ \hline
\# of valid cases & & 150 & 150 & &  \\ \hline
\end{tabular}%
}
\caption{RG experiment results. Best results per row in bold}
\end{table}
In the case of the RG problem, Table 3 shows that the advantage in processing time (avg) of \textbf{SD-DDF} with respect to SEQ is for all network sizes over 95\% getting over 98\% for 400 neurons. With respect to C-SEQ the advantage starts at 75.91\% for networks with 100 neurons and goes up to 87.69\% for networks with 1600 neurons. In general, \textbf{SD-DDF} is reaching similar results in the other variables except in those related to processing time where \textbf{SD-DDF} is much superior to SEQ and C-SEQ.\\
Figure 3 shows a representation of the time (avg) variable for the different dynamics and network sizes showing that for the network sizes considered, the trend of time (avg) grows polynomially in the three cases but with a significant difference in the coefficients of $x^2$: $2.28\cdot 10^{-5}$ (SEQ), $2.37\cdot 10^{-6}$ (C-SEQ) and $1.70\cdot 10^{-7}$ (\textbf{SD-DDF}).\\
%----------------
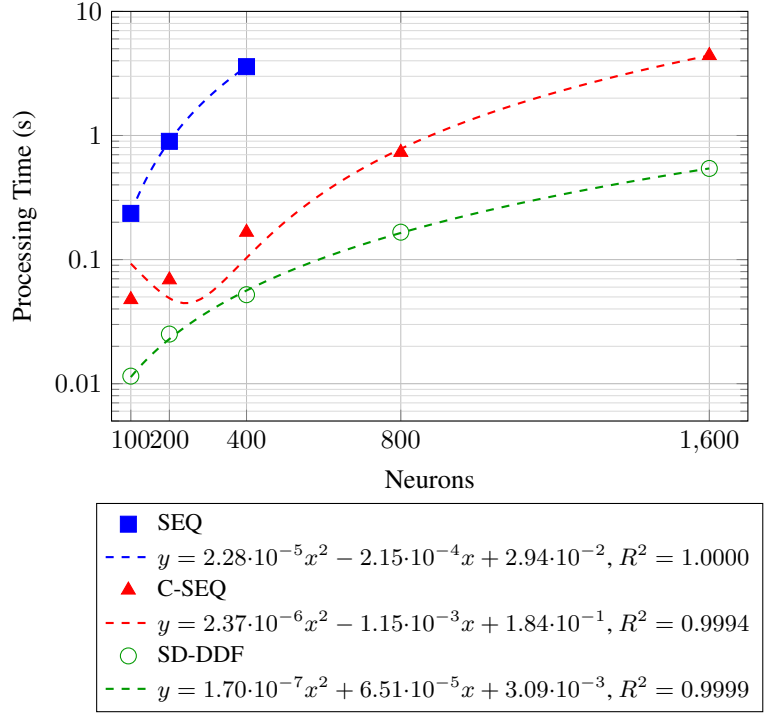
\begin{figure}[H]
    \centering
\begin{tikzpicture}
\begin{semilogyaxis}[
    xlabel={Neurons},
    ylabel={Processing Time (s)},
    legend style={
        font=\footnotesize,
        cells={anchor=west},
        at={(0.5,-0.2)},
        anchor=north,
        legend columns=1
    },
    grid=both,
    grid style={line width=0.1pt, draw=gray!30},
    major grid style={line width=0.2pt, draw=gray!50},
    width=10cm,
    height=7cm,
    xmin=50, xmax=1700,
    ymin=0.005, ymax=10,
    xtick={100, 200, 400, 800, 1600},
    log ticks with fixed point
]

% SEQ - Data points (only 100, 200, 400)
\addplot[only marks, mark=square*, mark size=3pt, blue]
    coordinates {(100,0.2357) (200,0.8975) (400,3.5876)};
\addlegendentry{SEQ}

% SEQ - Polynomial trendline
\addplot[blue, dashed, thick, domain=100:400, samples=100]
    {2.28e-05*x^2 - 2.15e-04*x + 2.94e-02};
\addlegendentry{$y = 2.28{\cdot}10^{-5}x^2 - 2.15{\cdot}10^{-4}x + 2.94{\cdot}10^{-2}$, $R^2=1.0000$}

% C-SEQ - Data points
\addplot[only marks, mark=triangle*, mark size=3pt, red]
    coordinates {(100,0.0476) (200,0.0687) (400,0.1661) (800,0.7315) (1600,4.4090)};
\addlegendentry{C-SEQ}

% C-SEQ - Polynomial trendline
\addplot[red, dashed, thick, domain=100:1600, samples=100]
    {2.37e-06*x^2 - 1.15e-03*x + 1.84e-01};
\addlegendentry{$y = 2.37{\cdot}10^{-6}x^2 - 1.15{\cdot}10^{-3}x + 1.84{\cdot}10^{-1}$, $R^2=0.9994$}

% SD-DDF - Data points
\addplot[only marks, mark=o, mark size=3pt, green!60!black]
    coordinates {(100,0.0115) (200,0.0251) (400,0.0521) (800,0.1663) (1600,0.5427)};
\addlegendentry{SD-DDF}

% SD-DDF - Polynomial trendline
\addplot[green!60!black, dashed, thick, domain=100:1600, samples=100]
    {1.70e-07*x^2 + 6.51e-05*x + 3.09e-03};
\addlegendentry{$y = 1.70{\cdot}10^{-7}x^2 + 6.51{\cdot}10^{-5}x + 3.09{\cdot}10^{-3}$, $R^2=0.9999$}

\end{semilogyaxis}
\end{tikzpicture}
\caption{RG Problem: Processing Time vs Network Size}
\label{fig:mi_grafica_tikz}
\end{figure}
%----------------
\subsection{N-Queen allocation (3-NQ)}
This classical problem consists of placing n queens on an $n$x$n$ chess board in such a way that the queens do not attack each other.\\
To model the problem, let $X=(x_{ij}),\ x_{ij}\in\{0,1\},\ i,j=1,...,n$ be the vector state with:
\[
x_{ij} =
\begin{cases}
    1, & \text{if there is one queen in row $i$ column $j$}\\
    0, & \text{otherwise}
\end{cases}
\]
So, the vector state $X$ has $n$x$n$ neurons. Three restrictions are given for this problem:
\begin{enumerate}
    \item One queen per row: $\sum_{j}x_{ij}=1,\ i=1,...,n$
    \item One queen per column: $\sum_{i}x_{ij}=1,\ j=1,...,n$
    \item Maximum one queen per diagonal: $\sum_{i,j\in diag}x_{ij}\leq 1$
\end{enumerate}
From the restrictions above and adding the Lagrange multipliers $(A, B, C)$, a first energy function can be expressed as follows:
\begin{align*} 
E=A·\sum_{i=1}^{n}(\sum_{j=1}^{n}x_{ij}-1)^2+B·\sum_{j=1}^{n}(\sum_{i=1}^{n}x_{ij}-1)^2+C·\sum_{diagonals}(\sum_{ij\in diag}x_{ij}-1)^2
\end{align*}
To express the model in a bipolar network with vector state $S(s_{ij}),\ s_{ij}\in \{-1,+1\},\ i,j=1,...,n$, we transform $x_{ij}=\frac{s_{ij}+1}{2}$ and operating with the first element of $E$:
\begin{align*}
&\sum_{j}x_{ij}=\sum_{j} \frac{s_{ij}+1}{2}=\frac{1}{2}\sum_{j}s_{ij}+\frac{n}{2} \\
&(\sum_{j}x_{ij}-1)=\frac{1}{2}\sum_{j}s_{ij}+\frac{n}{2}-1=\frac{1}{2}(\sum_{j}s_{ij}+(n-2)) \\
&(\sum_{j}x_{ij}-1)^2=\frac{1}{4}(\sum_{j}s_{ij}+(n-2))^2=\frac{1}{4}((\sum_{j}s_{ij})^2+(n-2)^2+2(n-2)\sum_{j}s_{ij})\\
&(\sum_{j}s_{ij})^2=\sum_{j}s_{jj}^2+2\sum_{k<l}s_{ik}s_{il}=n+2\sum_{k<l}s_{ik}s_{il}    
\end{align*}
\begin{align}
&A(\sum_{j}x_{ij}-1)^2=\frac{A}{4}(n+2\sum_{k<l}s_{ik}s_{il}+(n-2)^2+2(n-2)\sum_{j}s_{ij})= \nonumber \\
&=\frac{1}{4}(An+2A\sum_{k<l}s_{ik}s_{il}+A(n-2)^2+2(n-2)A\sum_{j}s_{ij})
\end{align}
The three terms of $E$ have the same mathematical form. 
By comparing equation $(18)$ with the general expression of the energy function of the Hopfield network:
\begin{align*}
    E=-\frac{1}{2}\sum_{ij}\sum_{kl}w_{ij,kl}s_{ij}s_{kl}+\sum_{ij}\theta_{ij}s_{ij}
\end{align*}
The contribution of the first term of $E$, corresponding to the restrictions per row to the variation of the energy function of the network, is: $-2A$ for quadratic terms $s_{ij}s_{kl}$ and $2(n-2)A$ for lineal terms $s_{ij}$. The rest of the elements in expression $(18)$ are constants and can be ignored as far as the variation of the energy function is concerned.\\
Similarly, with the multiplier $B$, the contribution of the second term of $E$, corresponding to the restrictions per column to the variation of the energy function of the network, is: $-2B$ for quadratic terms $s_{ij}s_{kl}$ and $2(n-2)B$ for lineal terms $s_{ij}$.\\
Analogously, the contribution of the third term of $E$, corresponding to the restrictions per diagonal to the variation of the energy function of the network, is: $-2C$ for the quadratic terms $s_{ij}s_{kl}$ and $2(L-2)C$ for the lineal terms $s_{ij}$, being $L$ the length of the diagonal in process.\\
To evaluate the quality of the solutions reached, in addition to the energy and processing time, the variable
Relative quality of the solution has been considered, calculated as:
\begin{align*}
nq\_cal\_rel = \frac{+|n-q|+Conflicts}{n}
\end{align*}
where $n$ is the number of queens to allocate, $q$ is the number of queens allocated and $Conflicts$ calculated as the number of queens attacking each other.
In the models for the NQ allocation experiments, values of $A=10$, $B=10$ and $C=3$ have been used.
\begin{table}[H]
\centering
\tiny
\resizebox{\textwidth}{!}{%
\renewcommand{\arraystretch}{0.8}
\begin{tabular}{|l|r|r|r|r|r|}
\hline
 & \multicolumn{3}{c|}{DYNAMICS} & \multicolumn{2}{c|}{VARIATION} \\ \hline
NEURONS & SEQ & C-SEQ & SD-DDF & SD-DDF/SEQ - 1 & SD-DDF/C-SEQ - 1 \\ \hline
\textbf{64 (8 queens)} & & & & & \\ \hline
energy (avg) & \textbf{-8067.63} & -8057.95 & -8066.69 & -0.0001 & 0.0011 \\ \hline
energy (min) & \textbf{-8168.00} & -8144.00 & \textbf{-8168.00} & 0.0000 & 0.0029 \\ \hline
energy (desvest) & 48.24 & \textbf{41.82} & 53.74 & 0.1141 & 0.2852 \\ \hline
time (avg) & 0.0131 & 0.0030 & \textbf{0.0016} & -0.8817 & -0.4802 \\ \hline
time (min) & 0.0057 & 0.0018 & \textbf{0.0011} & -0.8137 & -0.4117 \\ \hline
time (desvest) & 0.0051 & 0.0006 & \textbf{0.0003} & -0.9464 & -0.5504 \\ \hline
nq\_cal\_rel (mean) & 0.5158 & 0.5653 & \textbf{0.4750} & -0.0792 & -0.1597 \\ \hline
nq\_cal\_rel (min) & \textbf{0.0000} & 0.1250 & \textbf{0.0000} & & -1.0000 \\ \hline
nq\_cal\_rel (desvest) & 0.2794 & \textbf{0.2321} & 0.2550 & -0.0875 & 0.0985 \\ \hline
\# of valid cases & 150 & 113 & 150 & & \\ \hline
\textbf{100 (10 queens)} & & & & & \\ \hline
energy (avg) & -17147.17 & -17140.60 & \textbf{-17150.35} & 0.0002 & 0.0006 \\ \hline
energy (min) & \textbf{-17260.00} & \textbf{-17260.00} & \textbf{-17260.00} & 0.0000 & 0.0000 \\ \hline
energy (desvest) & 53.31 & 51.24 & \textbf{48.14} & -0.0971 & -0.0606 \\ \hline
time (avg) & 0.0603 & 0.0197 & \textbf{0.0051} & -0.9158 & -0.7419 \\ \hline
time (min) & 0.0208 & 0.0083 & \textbf{0.0019} & -0.9108 & -0.7760 \\ \hline
time (desvest) & 0.0270 & 0.0186 & \textbf{0.0028} & -0.8971 & -0.8509 \\ \hline
nq\_cal\_rel (mean) & \textbf{0.5660} & 0.6300 & 0.5865 & 0.0362 & -0.0691 \\ \hline
nq\_cal\_rel (min) & \textbf{0.1000} & \textbf{0.1000} & \textbf{0.1000} & 0.0000 & 0.0000 \\ \hline
nq\_cal\_rel (desvest) & \textbf{0.2046} & 0.2225 & 0.2335 & 0.1412 & 0.0493 \\ \hline
\# of valid cases & 150 & 120 & 148 & & \\ \hline
\textbf{400 (20 queens)} & & & & & \\ \hline
energy (avg) & \textbf{-162690.29} & -162664.22 & -162681.55 & -0.0001 & 0.0001 \\ \hline
energy (min) & \textbf{-162864.00} & -162840.00 & -162816.00 & -0.0003 & -0.0001 \\ \hline
energy (desvest) & 66.42 & 66.52 & \textbf{64.63} & -0.0269 & -0.0284 \\ \hline
time (avg) & 0.8055 & 0.1076 & \textbf{0.0302} & -0.9625 & -0.7190 \\ \hline
time (min) & 0.3794 & 0.0733 & \textbf{0.0115} & -0.9698 & -0.8437 \\ \hline
time (desvest) & 0.3306 & 0.0360 & \textbf{0.0335} & -0.8986 & -0.0682 \\ \hline
nq\_cal\_rel (mean) & \textbf{0.5933} & 0.6524 & 0.6123 & 0.0320 & -0.0614 \\ \hline
nq\_cal\_rel (min) & \textbf{0.2000} & 0.2500 & 0.3000 & 0.5000 & 0.2000 \\ \hline
nq\_cal\_rel (desvest) & 0.1595 & 0.1520 & \textbf{0.1515} & -0.0499 & -0.0035 \\ \hline
\# of valid cases & 150 & 126 & 150 & & \\ \hline
\textbf{900 (30 queens)} & & & & & \\ \hline
energy (avg) & & -580552.85 & \textbf{-580601.63} & & 0.0001 \\ \hline
energy (min) & & -580716.00 & \textbf{-580756.00} & & 0.0001 \\ \hline
energy (desvest) & & \textbf{69.77} & 75.19 & & 0.0778 \\ \hline
time (avg) & & 0.5280 & \textbf{0.0706} & & -0.8662 \\ \hline
time (min) & & 0.3879 & \textbf{0.0470} & & -0.8788 \\ \hline
time (desvest) & & 0.1581 & \textbf{0.0273} & & -0.8277 \\ \hline
nq\_cal\_rel (mean) & & 0.7060 & \textbf{0.6433} & & -0.0887 \\ \hline
nq\_cal\_rel (min) & & 0.4667 & \textbf{0.3667} & & -0.2143 \\ \hline
nq\_cal\_rel (desvest) & & \textbf{0.1148} & 0.1159 & & 0.0093 \\ \hline
\# of valid cases & & 123 & 150 & &  \\ \hline
\textbf{1600 (40 queens)} & & & & & \\ \hline
energy (avg) & & -1414873.84 & \textbf{-1414944.75} & & 0.0001 \\ \hline
energy (min) & & -1415088.00 & \textbf{-1415164.00} & & 0.0001 \\ \hline
energy (desvest) & & \textbf{91.91} & 94.27 & & 0.0256 \\ \hline
time (avg) & & 1.9208 & \textbf{0.2599} & & -0.8647 \\ \hline
time (min) & & 1.4411 & \textbf{0.1702} & & -0.8819 \\ \hline
time (desvest) & & 0.4272 & \textbf{0.1033} & & -0.7583 \\ \hline
nq\_cal\_rel (mean) & & 0.6948 & \textbf{0.6158} & & -0.1137 \\ \hline
nq\_cal\_rel (min) & & 0.4250 & \textbf{0.3500} & & -0.1765 \\ \hline
nq\_cal\_rel (desvest) & & \textbf{0.1103} & 0.1156 & & 0.0476 \\ \hline
\# of valid cases & & 111 & 150 & &  \\ \hline
\end{tabular}%
}
\caption{NQ experiment results. Best results per row in bold}
\end{table}
In the case of the NQ problem, Table 4 shows that the advantage in processing time (avg) of \textbf{SD-DDF} with respect to SEQ is for all network sizes over 88\% getting over 96\% for 400 neurons. With respect to C-SEQ the advantage starts at 48.02\% for networks with 100 neurons and goes up to 86.47\% for networks with 1600 neurons. In general, \textbf{SD-DDF} is reaching similar results in the other variables except in those related to processing time where \textbf{SD-DDF} is much superior to SEQ and C-SEQ.\\
Figure 4 shows a representation of the time (avg) variable for the different dynamics and network sizes showing that for the network sizes considered, the trend of time (avg) grows polynomially in the three cases but with a significant difference in the coefficient of $x^2$: $3.49\cdot 10^{-6}$ (SEQ), $8.66\cdot 10^{-7}$ (C-SEQ) and $1.09\cdot 10^{-7}$ (\textbf{SD-DDF}).\\
%------------
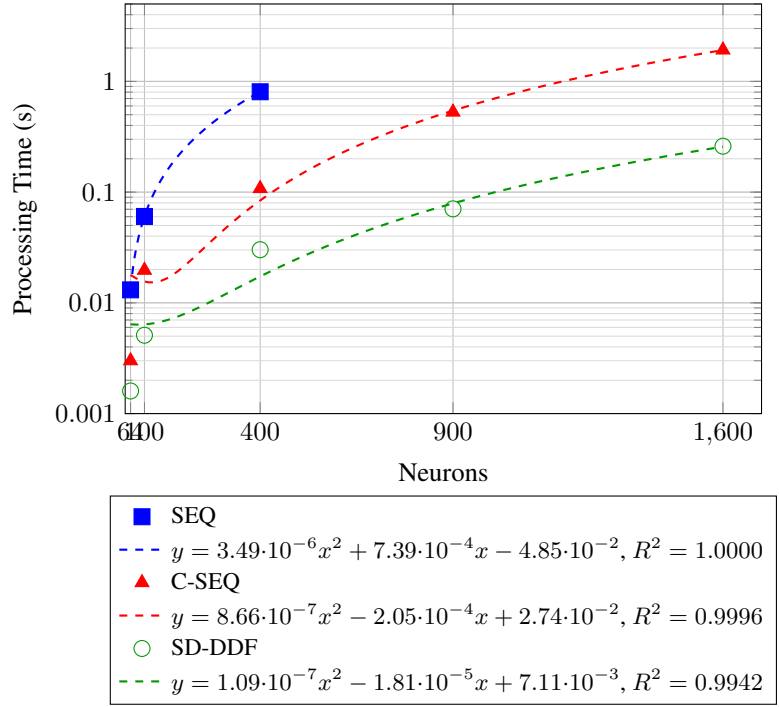
\begin{figure}[H]
    \centering
\begin{tikzpicture}
\begin{semilogyaxis}[
    xlabel={Neurons},
    ylabel={Processing Time (s)},
    legend style={
        font=\footnotesize,
        cells={anchor=west},
        at={(0.5,-0.2)},
        anchor=north,
        legend columns=1
    },
    grid=both,
    grid style={line width=0.1pt, draw=gray!30},
    major grid style={line width=0.2pt, draw=gray!50},
    width=10cm,
    height=7cm,
    xmin=50, xmax=1700,
    ymin=0.001, ymax=5,
    xtick={64, 100, 400, 900, 1600},
    log ticks with fixed point
]

% SEQ - Data points (only 64, 100, 400)
\addplot[only marks, mark=square*, mark size=3pt, blue]
    coordinates {(64,0.0131) (100,0.0603) (400,0.8055)};
\addlegendentry{SEQ}

% SEQ - Polynomial trendline
\addplot[blue, dashed, thick, domain=64:400, samples=100]
    {3.49e-06*x^2 + 7.39e-04*x - 4.85e-02};
\addlegendentry{$y = 3.49{\cdot}10^{-6}x^2 + 7.39{\cdot}10^{-4}x - 4.85{\cdot}10^{-2}$, $R^2=1.0000$}

% C-SEQ - Data points
\addplot[only marks, mark=triangle*, mark size=3pt, red]
    coordinates {(64,0.0030) (100,0.0197) (400,0.1076) (900,0.5280) (1600,1.9208)};
\addlegendentry{C-SEQ}

% C-SEQ - Polynomial trendline
\addplot[red, dashed, thick, domain=64:1600, samples=100]
    {8.66e-07*x^2 - 2.05e-04*x + 2.74e-02};
\addlegendentry{$y = 8.66{\cdot}10^{-7}x^2 - 2.05{\cdot}10^{-4}x + 2.74{\cdot}10^{-2}$, $R^2=0.9996$}

% SD-DDF - Data points
\addplot[only marks, mark=o, mark size=3pt, green!60!black]
    coordinates {(64,0.0016) (100,0.0051) (400,0.0302) (900,0.0706) (1600,0.2599)};
\addlegendentry{SD-DDF}

% SD-DDF - Polynomial trendline
\addplot[green!60!black, dashed, thick, domain=64:1600, samples=100]
    {1.09e-07*x^2 - 1.81e-05*x + 7.11e-03};
\addlegendentry{$y = 1.09{\cdot}10^{-7}x^2 - 1.81{\cdot}10^{-5}x + 7.11{\cdot}10^{-3}$, $R^2=0.9942$}

\end{semilogyaxis}
\end{tikzpicture}
\caption{NQ Problem: Processing Time vs Network Size}
\label{fig:mi_grafica_tikz}
\end{figure}
%------------
\subsection{Traveling Salesman Problem (4-TSP)}
The Traveling Salesman Problem is one of the most popular and well studied problems in the literature of combinatorial optimization and is the first combinatorial problem treated by a Hopfield Network (Hopfield and Tank, 1985). The problem is NP-complete (Garey and Johnson, 1979).
The problem consists of defining, for a given set of $n$ cities, a circular tour that visits each city exactly once and has minimal length.
To represent the problems in terms of a Hopfield Network, a matrix of $n^2$ neurons will be used: each row represents a city and each column represents a place in the tour. And so:
\[
v_{it} =
\begin{cases}
    1, & \text{if city i is visited in place t of the tour}\\
    0, & \text{otherwise}
\end{cases}
\]
To work with bipolar neurons, the following representation will be used:
\[
s_{it} =
\begin{cases}
    +1, & \text{if city i is visited in place t of the tour}\\
    -1, & \text{otherwise}
\end{cases}
\]
And the equivalence is given by: $v_{it}=\frac{s_{it}+1}{2}$. \
Restrictions $(1 \& 2)$ and the objective function $(3)$ are given for the definition of the energy function:
\begin{enumerate}
    \item One city in each position on the tour: $\sum_{i=1}^{n}v_{it}=1, \  \forall t=1,2,...,n$.\\
    In terms of bipolar neurons: $\sum_{i=1}^{n}s_{it}=2-n, \forall t=1,2,...,n$
    \item Each city is visited only once: $\sum_{t=1}^{n}v_{it}=1, \ \forall i=1,2,...,n$.\\
    In terms of bipolar neurons: $\sum_{t=1}^{n}s_{it}=2-n, \forall i=1,2,...n$
    \item Minimal length of the tour: $\sum_{t=1}^{n}\sum_{i=1}^{n}\sum_{j=1}^{n}d_{ij}v_{it}v_{j,t+1}$ with index t calculated module n to allow circularity in the tour.\\
    \\In terms of bipolar neurons: $\sum_{i=1}^{n}\sum_{j=1}^{n}\sum_{t=1}^{n}\frac{1}{4}d_{ij}(s_{it}s_{j,t+1}+s_{it}+s_{j,t+1}+1)$
\end{enumerate}
$d_{ij}$ is the distance between city $i$ and city $j$ that has: $d_{ij}=d_{ji}$ and $d_{ii}=0$. \\

The general form of the energy function of the network is:
\begin{align*}
    E=-\frac{1}{2}\sum_{it}\sum_{jr}w_{it,jr}s_{it}s_{jr}+\sum_{it}\theta_{it}s_{it}
\end{align*}
Given the constraints and the objective function to be minimized described above, the energy function can be expressed as follows
\begin{align*}
    E = A·E_{1}+B·E_{2}+E_{3}
\end{align*}
where $A$ and $B$ are the Lagrange coefficients. $E_{1}, \, E_{2}$ and $E_{3}$ come from the above restrictions: 
 
\begin{enumerate}
    \item $E_{1}$ One city in each position on the tour
    \begin{align*}
    &E_{1}=\sum_{t=1}^{n}(\sum_{i=1}^{n}s_{it}-(2-n))^2=\sum_{t=1}^{n}((\sum_{i=1}^{n}s_{it})^2-2(2-n)\sum_{i=1}^{n}s_{it}+(2-n)^2)\\
    \text{given that}\\
    &(\sum_{i=1}^{n}
    s_{it})^2=\sum_{i=1}^{n}s_{ii}^2+\sum_{i=1}^{n}\sum_{j\neq i}s_{it}s_{jt}=n+\sum_{i=1}^{n}\sum_{j\neq i}s_{it}s_{jt}\\
    \text{we obtain}\\
    &AE_{1}=\sum_{t=1}^{n}(\sum_{i=1}^{n}\sum_{j\neq i}As_{it}s_{jt}+\sum_{i=1}^{n}A2(n-2)s_{it})+Cte
    \end{align*}
    \item $E_{2}$ Each city is visited only once. Analogously:
    \begin{align*}
    &BE_{2}=\sum_{i=1}^{n}(\sum_{t=1}^{n}\sum_{t\neq r}Bs_{it}s_{ir}+\sum_{t=1}^{n}B2(n-2)s_{it})+Cte
    \end{align*}
    \item $E_{3}$ Minimal length of the tour
    \begin{align*}
    &E_{3}=\sum_{t=1}^{n}\sum_{i=1}^{n}\sum_{j\neq i}\frac{1}{4}d_{ij}(s_{it}s_{j,t+1}+s_{it}+s_{j,t+1}+1)=\\
    &=\sum_{t=1}^{n}\sum_{i=1}^{n}\sum_{j\neq i}\frac{1}{4}d_{ij}s_{it}s_{j,t+1}+\sum_{t=1}^{n}\sum_{i=1}^{n}\sum_{j\neq i}\frac{1}{2}d_{ij}s_{it}+Cte
    \end{align*}
\end{enumerate}
By comparing the quadratic and linear components of the two expressions of the energy function of the network given above, the contribution of each restriction to the coefficients $w$ and threshold values $\theta$ is:
\[
w_{it,jr}=
\begin{cases}
    -2A \ when \ t=r \ and \ j\neq i\\
    -2B \ when \ t\neq r \ and \ i=j\\
    -\frac{1}{2}d_{ij} \ when \ i\neq j \ and \ r=t+1(mod\ n)\\
    0 \ otherwise
\end{cases}
\]
\[
\theta_{it}=2A(n-2)+2B(n-2)+\frac{1}{2}\sum_{i=1}^{n}\sum_{j\neq i}d_{ij}
\]
In the experiment, we used parameters of $A = B = 20$. Inter-city distances were randomly generated within the range [1, 100].\\
\\
In the case of the TSP problem, Table 5 shows the data for the valid tours generated. The advantage in processing time (avg) of \textbf{SD-DDF} with respect to SEQ is for all network sizes over 89\% getting over 96\% for 400 neurons. With respect to C-SEQ the advantage starts at 54.32\% for networks with 100 neurons and goes up to 85.37\% for networks with 1600 neurons. In general, \textbf{SD-DDF} is reaching similar results in the other variables except in those related to processing time where \textbf{SD-DDF} is much superior to SEQ and C-SEQ.\\
Figure 5 shows a representation of the time (avg) variable for the different dynamics and network sizes showing that for the network sizes considered, the trend of time (avg) grows polynomially in the three cases, but with a significant difference in the coefficient of $x^2$: $1.25\cdot 10^{-6}$ (SEQ), $8.45\cdot 10^{-7}$ (C-SEQ) and $1.34\cdot 10^{-7}$ (\textbf{SD-DDF}).\\
\begin{table}[H]
\centering
\tiny
\resizebox{\textwidth}{!}{%
\renewcommand{\arraystretch}{0.8}
\begin{tabular}{|l|r|r|r|r|r|}
\hline
 & \multicolumn{3}{c|}{DYNAMICS} & \multicolumn{2}{c|}{VARIATION} \\ \hline
NEURONS & SEQ & C-SEQ & SD-DDF & SD-DDF/SEQ - 1 & SD-DDF/C-SEQ - 1 \\ \hline
\textbf{64 (8 cities)} & & & & & \\ \hline
energy (avg) & \textbf{-19319.38} & -19283.91 & -19299.34 & -0.0010 & 0.0008 \\ \hline
energy (min) & \textbf{-19656.00} & -19679.00 & -19659.00 & 0.0002 & -0.0010 \\ \hline
energy (desvest) & \textbf{337.85} & 354.77 & 349.77 & 0.0353 & -0.0141 \\ \hline
time (avg) & 0.0133 & 0.0031 & \textbf{0.0014} & -0.8920 & -0.5432 \\ \hline
time (min) & 0.0058 & 0.0019 & \textbf{0.0010} & -0.8233 & -0.4647 \\ \hline
time (desvest) & 0.0057 & 0.0006 & \textbf{0.0002} & -0.9584 & -0.5700 \\ \hline
tour length (mean) & \textbf{382.93} & 391.66 & 401.42 & 0.0483 & 0.0249 \\ \hline
tour length (min) & \textbf{226.00} & 238.00 & 249.00 & 0.1018 & 0.0462 \\ \hline
tour length (desvest) & 56.07 & 50.61 & \textbf{48.29} & -0.1388 & -0.0458 \\ \hline
\# of valid cases & 144 & 130 & 143 & & \\ \hline
\textbf{100 (10 cities)} & & & & & \\ \hline
energy (avg) & \textbf{-40189.24} & -40215.90 & -40214.48 & 0.0006 & 0.0000 \\ \hline
energy (min) & \textbf{-40800.00} & -40822.00 & -40857.00 & 0.0014 & 0.0009 \\ \hline
energy (desvest) & 377.52 & \textbf{367.22} & 379.01 & 0.0039 & 0.0321 \\ \hline
time (avg) & 0.0938 & 0.0193 & \textbf{0.0048} & -0.9492 & -0.7527 \\ \hline
time (min) & 0.0281 & 0.0084 & \textbf{0.0030} & -0.8920 & -0.6405 \\ \hline
time (desvest) & 0.0772 & 0.0152 & \textbf{0.0028} & -0.9642 & -0.8176 \\ \hline
tour length (mean) & 476.88 & \textbf{464.21} & 470.73 & -0.0129 & 0.0140 \\ \hline
tour length (min) & 315.00 & 358.00 & \textbf{303.00} & -0.0381 & -0.1536 \\ \hline
tour length (desvest) & 57.40 & 55.12 & \textbf{54.74} & -0.0463 & -0.0069 \\ \hline
\# of valid cases & 130 & 135 & 143 & & \\ \hline
\textbf{400 (20 cities)} & & & & & \\ \hline
energy (avg) & -368949.05 & -368769.65 & \textbf{-369064.83} & 0.0003 & 0.0008 \\ \hline
energy (min) & \textbf{-372899.00} & -372897.00 & -372864.00 & -0.0001 & -0.0001 \\ \hline
energy (desvest) & 3563.87 & 3580.48 & \textbf{3515.76} & -0.0135 & -0.0181 \\ \hline
time (avg) & 0.8903 & 0.1105 & \textbf{0.0279} & -0.9687 & -0.7478 \\ \hline
time (min) & 0.4246 & 0.0704 & \textbf{0.0135} & -0.9681 & -0.8077 \\ \hline
time (desvest) & 0.3289 & 0.0375 & \textbf{0.0238} & -0.9275 & -0.3633 \\ \hline
tour length (mean) & 941.84 & \textbf{941.55} & 955.17 & 0.0142 & 0.0145 \\ \hline
tour length (min) & \textbf{670.00} & 711.00 & 707.00 & 0.0552 & -0.0056 \\ \hline
tour length (desvest) & 88.20 & \textbf{82.42} & 83.94 & -0.0483 & 0.0185 \\ \hline
\# of valid cases & 135 & 125 & 135 & & \\ \hline
\textbf{900 (30 cities)} & & & & & \\ \hline
energy (avg) & & \textbf{-1302651.74} & -1302455.92 & & -0.0002 \\ \hline
energy (min) & & \textbf{-1310054.00} & -1309969.00 & & -0.0001 \\ \hline
energy (desvest) & & 5543.73 & \textbf{5510.51} & & -0.0060 \\ \hline
time (avg) & & 0.5976 & \textbf{0.0742} & & -0.8758 \\ \hline
time (min) & & 0.3782 & \textbf{0.0468} & & -0.8763 \\ \hline
time (desvest) & & 0.2241 & \textbf{0.0310} & & -0.8617 \\ \hline
tour length (mean) & & \textbf{1399.57} & 1462.43 & & 0.0449 \\ \hline
tour length (min) & & \textbf{1172.00} & 1182.00 & & 0.0085 \\ \hline
tour length (desvest) & & \textbf{93.04} & 98.78 & & 0.0617 \\ \hline
\# of valid cases & & 138 & 136 & & \\ \hline
\textbf{1600 (40 cities)} & & & & & \\ \hline
energy (avg) & & -3145238.49 & \textbf{-3145453.16} & & 0.0001 \\ \hline
energy (min) & & \textbf{-3160135.00} & -3160125.00 & & 0.0000 \\ \hline
energy (desvest) & & 11347.84 & \textbf{11327.97} & & -0.0018 \\ \hline
time (avg) & & 2.0043 & \textbf{0.2932} & & -0.8537 \\ \hline
time (min) & & 1.3920 & \textbf{0.1710} & & -0.8772 \\ \hline
time (desvest) & & 0.4588 & \textbf{0.1732} & & -0.6225 \\ \hline
tour length (mean) & & \textbf{1833.51} & 1873.65 & & 0.0219 \\ \hline
tour length (min) & & \textbf{1515.00} & 1653.00 & & 0.0911 \\ \hline
tour length (desvest) & & 121.59 & \textbf{102.81} & & -0.1544 \\ \hline
\# of valid cases & & 130 & 144 & & \\ \hline
\end{tabular}%
}
\caption{TSP experiment results. Best results per row in bold}
\end{table}
%-------------
\begin{figure}[H]
\centering
\begin{tikzpicture}
\begin{semilogyaxis}[
    xlabel={Neurons},
    ylabel={Processing Time (s)},
    grid=major,
    width=12cm,
    height=8cm,
    legend style={
        at={(0.5,-0.25)},
        anchor=north,
        legend columns=1,
        font=\footnotesize,
        cells={anchor=west}
    },
    xmin=0, xmax=1700,
    ymin=0.001, ymax=10,
    xtick={0, 200, 400, 600, 800, 1000, 1200, 1400, 1600},
]

% SEQ data points (blue squares)
\addplot[
    color=blue,
    mark=square*,
    mark size=3pt,
    thick,
    only marks
] coordinates {
    (64,0.0133) (100,0.0938) (400,0.8903)
};

% SEQ polynomial trendline
\addplot[
    color=blue,
    dashed,
    thick,
    domain=64:400,
    samples=100
] {1.25e-6*x^2 + 2.03e-3*x - 1.22e-1};
\addlegendentry{SEQ: $y = 1.25 \cdot 10^{-6}x^2 + 2.03 \cdot 10^{-3}x - 0.122$, $R^2 = 1.0000$}

% C-SEQ data points (red triangles)
\addplot[
    color=red,
    mark=triangle*,
    mark size=3pt,
    thick,
    only marks
] coordinates {
    (64,0.0031) (100,0.0193) (400,0.1105) (900,0.5976) (1600,2.0043)
};

% C-SEQ polynomial trendline
\addplot[
    color=red,
    dashed,
    thick,
    domain=64:1600,
    samples=100
] {8.45e-7*x^2 - 1.09e-4*x + 1.49e-2};
\addlegendentry{C-SEQ: $y = 8.45 \cdot 10^{-7}x^2 - 1.09 \cdot 10^{-4}x + 0.015$, $R^2 = 0.9999$}

% SD-DDF data points (green circles)
\addplot[
    color=green!60!black,
    mark=*,
    mark size=3pt,
    thick,
    only marks
] coordinates {
    (64,0.0014) (100,0.0048) (400,0.0279) (900,0.0742) (1600,0.2932)
};

% SD-DDF polynomial trendline
\addplot[
    color=green!60!black,
    dashed,
    thick,
    domain=64:1600,
    samples=100
] {1.34e-7*x^2 - 3.73e-5*x + 8.31e-3};
\addlegendentry{SD-DDF: $y = 1.34 \cdot 10^{-7}x^2 - 3.73 \cdot 10^{-5}x + 0.008$, $R^2 = 0.9953$}

\end{semilogyaxis}
\end{tikzpicture}
\caption{TSP Problem: Processing Time vs Network Size}
\label{fig:tsp_dynamics}
\end{figure}
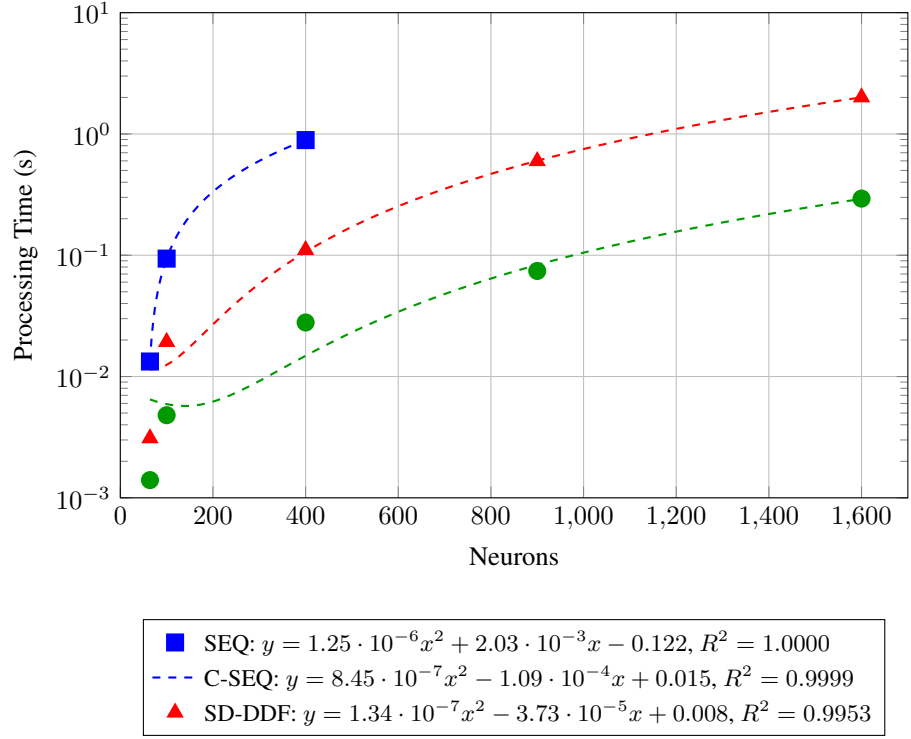
%-------------
\subsection{Overall comparison of SEQ, C-SEQ and \textbf{SD-DDF}}
A statistical analysis has been performed to compare the results obtained by the three algorithms under the following scheme:
\begin{enumerate}
    \item Descriptive statistics
    \item Normality: Shapiro-Wilk per (algorithm, instance) group
    \item Global test: Friedman (non-parametric) or one-way ANOVA (parametric)
    \item Post-hoc pairwise: Wilcoxon + Holm correction  |  Tukey HSD 
    \item Effect size: Vargha-Delaney A12  |  Cohen's d
\end{enumerate}
The problems analyzed are BP, RD (for 100, 200, 400 neurons) and NQ, TSP (for 64, 100, 400 neurons, corresponding to 8, 10, 20 queens/cities). The summary of results is:
\begin{enumerate}
    \item \textbf{Variable: energy}
    \begin{enumerate}
        \item All three algorithms reach average values of energy quite similar in all scenarios.
        \item In BP, C-SEQ reach the best values.
        \item In RD, C-SEQ and SEQ reaches the best values
        \item In NQ, \textbf{SD-DDF} and SEQ reaches the best values.
        \item In TSP, all three reach quite similar values
        \item \textbf{Final Verdict for variable energy:} Despite the slight numerical differences observed, statistical tests conclude in all scenarios that no significative differences have been detected in pair based comparisons among SEQ, C-SEQ and \textbf{SD-DDF}. In conclusion, statistically SEQ, C-SEQ and \textbf{SD-DDF} achieves the same energy values.
    \end{enumerate}
    \item \textbf{Variable: processing time}
    \begin{enumerate}
        \item \textbf{SD-DDF} is systematically the fastest algorithm in all scenarios.
        \item C-SEQ consistently ranks second in processing time.
        \item SEQ always turns out to be the slowest of the three algorithms, and the difference increases as the network size increases.
        \item \textbf{Final Verdict for variable time:} In post-hoc pairwise comparisons, the empirical effect size for the processing time was repetitively "large" demonstrating the superiority in processing time of \textbf{SD-DDF} with respect to C-SEQ and the superiority of C-SEQ with respect to SEQ.
    \end{enumerate}
    \item \textbf{Overall conclusion}
    \begin{enumerate}
        \item There is a tie in the energy score since none of the three algorithms analyzed manages to outperform the other two in a statistically significant way
        \item \textbf{SD-DDF} is the clear winner regarding processing time, reaching the best values in all scenarios.
        \item \textbf{Globally, \textbf{SD-DDF} is superior computationally to C-SEQ and SEQ, offering the fastest convergence without compromising the values of energy obtained. C-SEQ is the second and the original SEQ is the worst due to its excessive slowness.}
    \end{enumerate}
\end{enumerate}

\section{Discussions and Conclusions}
The new \textbf{Discrete Differential Filter (DDF)} introduced in this paper provides a general tool for HNNs (Hopfield Neural Networks, bipolar, discrete time, symmetric and zero diagonal weight matrix $W$) that allow to select from any subset $C$ of neurons of $R$ that would individually change their state at time $t$ according to asynchronous dynamics, a subset $C'\subset C$ that guarantees that, by changing in $R$ the state of all neurons in $C'$ synchronously at time $t$, the energy of the network $R$ will decrease or remain the same and the decrease will be a maximum. And, to achieve this, the \textbf{DDF} uses HNNs to solve HNNs.\\
In addition to its other applications, the new \textbf{DDF} serves as the basis for the definition of a new synchronous dynamics which we called \textbf{SD-DDF (Synchronous dynamics based upon Discrete Differential Filter)} which provides a solution for $R$ in less processing time than asynchronous dynamics.\\
\\
In the four experiments carried out, SD-DDF has achieved a significant reduction in processing time with respect to C-SEQ and to SEQ while similar values for the other variables were obtained. Moreover, in all cases, the values of time related variables (average, minimum and standard deviation) have been the best in each row, empirically demonstrating the superior performance in processing time of the new dynamics based on the concept of the \textbf{DDF} of $R$ at $S(t)$.\\
\\
While the processing time has been drastically reduced, the same cannot be said for the other variables studied. It would be desirable for the filter to also produce better data, particularly regarding the energy output. Future research should focus on both the filter and the synchronous dynamics to improve the quality of the solutions obtained. Subsequent work should address, among others, issues such as:
\begin{enumerate}
    \item Review classical HNNs with the new \textbf{SD-DDF}: how can the solution to problems that have been treated with asynchronous dynamics be improved.
    \item Use of the \textbf{DDF} to deal with large HNNs: similarly to how hardware parallelization works, \textbf{DDF} can help to process a large HNNs in a number of smaller HNNs.
    \item Apply simulated annealing to \textbf{SD-DDF}: including stochasticity in both: the selection of $C$ and the solution of each \textbf{DDF} at each time step, may improve the quality of the solutions reached, and not only the processing time.  
    \item Apply hysteresis in \textbf{DDF}: the hysteresis of the neurons has been proved to make the neurons more resistant to noisy inputs and more firmly hold the memory states.
    \item Apply \textbf{DDF} to multi-valued networks.
    \item Apply self-modelling in \textbf{DDF} to avoid local minima.
    \item Apply these ideas to Modern Hopfield Networks to allow the obtention of several tokens in each access.
\end{enumerate}

\bibliographystyle{plain}
\bibliography{biblio} 

\end{document}